\documentclass[11pt]{article}

\usepackage[preprint]{acl}

\usepackage{times}
\usepackage{latexsym}

\usepackage[T1]{fontenc}

\usepackage[utf8]{inputenc}

\usepackage{microtype}

\usepackage{inconsolata}

\usepackage{graphicx}
\usepackage{booktabs}
\usepackage{amsmath,amssymb}
\usepackage{tikz}
\usetikzlibrary{positioning,arrows.meta,fit,calc}
\usepackage{tikz}
\usetikzlibrary{arrows.meta,shapes.geometric}
\title{Jailbreak Detection on Large Language Models via Contrastive Prototype Structures}

\author{
  Yan Qiu \\
  Department of Mathematics \\
  Technical University of Munich \\
  Garching, Germany \\
  \texttt{go56xol@mytum.de}
  \And
  Yuanyuan Li \\
  Insure AI \\
  Munich Re \\
  San Francisco, USA \\
  \texttt{yli@munichre.com}
}

\begin{document}
\maketitle
\begin{abstract}
Jailbreak attempts and harmful prompts remain a challenge for the safe deployment of large language models. We study whether contrastively learned representations can support safety detection without training an additional downstream classifier. We propose Jailbreak Detection with Prototype Inference (JDPI), which uses binary safety labels to construct harmful and benign prototypes and makes predictions from relative prototype distances. JDPI uses only binary safety labels for encoder adaptation, prototype construction, and validation. In our setup, it uses fewer labeled examples than the classifier baseline (Method~A) by avoiding the separate split used to train the downstream classifier.

Across four benchmarks covering different input formats, contrastive fine-tuning improves JDPI's detection accuracy by 9.1--18.0 percentage points. Using the same encoder, JDPI achieves slightly higher ROC-AUC on one benchmark and stays within 0.017 of the baseline across all four. With dataset-specific adaptation, JDPI exceeds the released Llama Guard 3-8B baseline by 0.173--0.310 in ROC-AUC across the four benchmarks. In multi-turn replay on TomGibbs, JDPI uses accumulated user history as the conversation unfolds, achieving higher conversation-level accuracy than when it uses the current turn alone. Across 12 directed transfers, JDPI outperforms the classifier by an average of 1.59 percentage points in accuracy, demonstrating stronger cross-dataset robustness.
\end{abstract}

\section{Introduction}

For cybersecurity and safety, AI agents are typically deployed with policy constraints and safety guardrails. However, the increasing use of external tools and environments also introduces emerging security risks, including indirect prompt injection \cite{zhan-etal-2024-injecagent}, tool-integrated agent security \cite{NEURIPS2024_97091a51}, and broader agent attack and defense scenarios \cite{zhang2025agentsecuritybenchasb}. Despite these safeguards, security vulnerabilities and failure modes remain \cite{perez2022redteaminglanguagemodels, chao2024jailbreakbenchopenrobustnessbenchmark, mazeika2024harmbenchstandardizedevaluationframework, ganguli2022redteaminglanguagemodels} and so-called "jailbreaking" techniques are also commonly used to bypass safety alignment \cite{wei2023jailbrokendoesllmsafety, zou2023universaltransferableadversarialattacks}, which can pose risks to current AI-agent applications, starting from LLM use cases.

Many safety guards have been constructed at different levels. One commonly used tool is a classifier, which usually formulates safety detection as a classification problem \cite{inan2023llamaguardllmbasedinputoutput, sharma2025constitutionalclassifiersdefendinguniversal}, and the workflow can be: first define or construct a series of policy-violating cases, and then train the classifier to detect whether an input or output is intended to trigger harmful content.

The problem arises from this setup itself: first, the taxonomies are usually designed according to predefined safety policies or risk categories \cite{mazeika2024harmbenchstandardizedevaluationframework, markov2023holisticapproachundesiredcontent} and may not fully cover all cases. Second, attention may be directed toward distinguishing detailed categories rather than the underlying intent of the context itself. Third, the boundaries between detailed categories can be fuzzy and may not correspond directly to the structure of the representation space \cite{inan2023llamaguardllmbasedinputoutput, sharma2025constitutionalclassifiersdefendinguniversal, galinkin2024improvedlargelanguagemodel}. This motivates the question of whether harmful and benign inputs also form exploitable geometric structures in the learned representation space. If such structure exists, it raises two related questions: whether representation-aware structural features can improve conventional classifiers, and whether the classification process can be replaced by direct geometric inference based on the learned representation geometry.

We propose Jailbreak Detection with Prototype Inference (JDPI). We use contrastive learning to improve the separation between harmful and benign inputs in the representation space\cite{gao2022simcsesimplecontrastivelearning, khosla2021supervisedcontrastivelearning, chen2020simpleframeworkcontrastivelearning}. Using binary safety labels, JDPI constructs harmful and benign prototypes and classifies inputs by their relative distances to
these prototypes. This design eliminates the need for a separate labeled split to train a downstream classifier.

Methods B--D test whether adding explicit clustering-derived features improves classifier-based detection. Because the encoder is adapted to labeled data from a particular
source, we test cross-dataset robustness by transferring the detector to other datasets without target-dataset adaptation. Our four benchmarks cover prompt-level, question--answer, and multi-turn inputs and include adversarial examples. We assess whether retaining user history improves multi-turn detection on TomGibbs.

\section{Related Work}
\label{gen_inst}
A wide range of defense mechanisms has been proposed to improve LLM safety. Some methods enhance safety by prioritizing higher-level instructions over lower-level instructions during inference \cite{wallace2024instructionhierarchytrainingllms}, while others train models to reason about explicit safety specifications before generating responses \cite{guan2025deliberativealignmentreasoningenables}.  Robustness has also been improved through adversarial training and the incorporation of adversarial objectives \cite{sheshadri2025latentadversarialtrainingimproves}. Additional approaches intervene directly at the representation or activation level through mechanisms such as representation rerouting \cite{zou2024improvingalignmentrobustnesscircuit}. Traditional defense strategies, including input detection, preprocessing, and adversarial training, have also been systematically evaluated \cite{jain2023baselinedefensesadversarialattacks}. Input perturbation methods have also been explored, such as SmoothLLM \cite{robey2024smoothllmdefendinglargelanguage}. Other work investigates internal hidden-layer representations and performs layer-wise safety analysis followed by binary classification \cite{kadali2026jailbreakingleavestraceunderstanding}. Classifier-based detection is also widely used in safety guard systems \cite{inan2023llamaguardllmbasedinputoutput, sharma2025constitutionalclassifiersdefendinguniversal}. Embedding-based jailbreak detectors have combined pretrained sentence representations with conventional classifiers \cite{galinkin2024improvedlargelanguagemodel}. Despite these advances, many widely used guard models are constructed around predefined safety policies, risk categories, or labeled safety datasets. However, generalization to emerging attack strategies and previously unseen harmful behaviors remains a challenge. Furthermore, the opposite failure mode, namely over-refusal of benign inputs, has received increasing attention and has been studied explicitly in recent work \cite{cui2025orbenchoverrefusalbenchmarklarge}.

To determine whether latent representations contain meaningful safety signals, several studies have investigated safety-related information encoded within internal model states \cite{ball2024understandingjailbreaksuccessstudy, jin2024jailbreakhuntervisualanalyticsapproach}. JBShield identifies activated safety-related concepts in internal representations and leverages them for jailbreak detection and mitigation \cite{zhang2025jbshielddefendinglargelanguage}. SIREN further exploits internal safety-relevant features distributed across model layers for harmful-content detection \cite{jiao-etal-2026-llm}. RCS similarly performs jailbreak detection through representational contrastive scoring in latent space \cite{hua-etal-2026-rethinking}. Collectively, these studies suggest that internal representations contain rich geometric and semantic information that can be effectively leveraged for safety assessment.

A growing number of benchmarks evaluate jailbreak and safety-related behavior using large collections of adversarial and red-teaming-derived inputs, covering both harmful and benign requests. Many of these benchmarks provide comprehensive collections of both harmful and benign prompts, \citet{shen2024donowcharacterizingevaluating} collect and analyze in-the-wild jailbreak prompts, characterizing their attack strategies and evaluating their effectiveness against LLMs and external safeguards. WildJailbreak and WildGuardMix contain vanilla and adversarial prompts covering harmful and benign requests, including synthetic adversarial prompts generated using tactics mined from in-the-wild interactions \cite{jiang2024wildteamingscaleinthewildjailbreaks, wildguard2024}. BeaverTails provides question--answer pairs derived from red-teaming-related prompts designed to elicit unsafe responses, with safety annotations defined over the complete QA pair \cite{ji2023beavertailsimprovedsafetyalignment}. \citet{gibbs2024emergingvulnerabilitiesfrontiermodels} introduce paired single-turn and multi-turn cipher-based jailbreak inputs, alongside benign controls, and examine how prompting structure and conversational context affect guardrail performance. Evaluation frameworks such as StrongREJECT further emphasize that jailbreak assessment should account for whether harmful behavior is actually elicited rather than only whether an attack prompt is present \cite{souly2024strongrejectjailbreaks}.

Contrastive learning \cite{chen2020simpleframeworkcontrastivelearning} improves representation structure by increasing the separation between semantically different samples. This principle has been applied to sentence representation learning \cite{gao2022simcsesimplecontrastivelearning} and supervised class-aware representation learning \cite{khosla2021supervisedcontrastivelearning}. Sentence-transformer models provide fixed-dimensional sentence representations that are suitable for efficient similarity and distance computation \cite{reimers2019sentencebertsentenceembeddingsusing}, extending pretrained Transformer encoders such as BERT \cite{devlin2019bertpretrainingdeepbidirectional}. In our work, an MPNet-based sentence-transformer encoder is used to obtain semantic representations for subsequent structure analysis \cite{song2020mpnetmaskedpermutedpretraining}.

Prototypical Networks introduced prototype-based few-shot classification, where samples are classified according to distances from representative class prototypes in a learned metric space \cite{snell2017prototypicalnetworksfewshotlearning}. Prototype-oriented Conditional Transport \cite{tanwisuth2021prototypeorientedframeworkunsuperviseddomain} addresses source--target alignment by aligning target features with class prototypes. Recent safety detection methods have similarly explored geometric signals in representation space. RCS distinguishes benign and malicious inputs through contrastive scoring over internal model representations \cite{hua-etal-2026-rethinking}, while T3 models the distribution of acceptable prompts in semantic space and formulates safety detection as out-of-distribution detection without requiring harmful training examples \cite{ganguly2026trusttypical}. The Semantic Codebooks approach compares multilingual query embeddings against a fixed codebook of jailbreak prompts, providing a training-free external guardrail based on semantic similarity 
\cite{alanova2026crosslingualjailbreakdetectionsemantic}.

Different from these approaches, our focus is not to construct a stronger embedding model or to model only a predefined safe or malicious reference distribution. Instead, we investigate whether geometric structure learned under binary harmful benign supervision can reduce the dependence on an additional downstream classifier. Using the same contrastively adapted encoder, we compare embedding-based classification and variants augmented with clustering-derived features against direct inference from harmful and benign prototype sets, without requiring fine-grained risk annotations. But how finer cluster structure and its refinement affect binary harmful--benign detection is still examined.

\section{Method}
\label{headings}

\begin{figure*}[t]
\centering

\begin{tikzpicture}[
    font=\small,
    >=Latex,
    node distance=7mm and 10mm,
    box/.style={
        draw,
        rectangle,
        align=center,
        text width=2.4cm,
        minimum height=9mm,
        inner xsep=3pt,
        inner ysep=4pt
    },
    method/.style={
        draw,
        rectangle,
        align=center,
        text width=1.95cm,
        minimum height=9mm,
        inner xsep=3pt,
        inner ysep=4pt
    },
    arrow/.style={
        ->,
        line width=0.7pt
    }
]


\node[box] (data)
{Inputs + binary labels};

\node[box, text width=5.4cm, right=of data] (contrastive)
{Contrastive representation learning\\
Phase 1: binary separation};

\node[box, right=of contrastive] (embedding)
{Fine-tuned encoder\\
$\mathbf{z}=f_{\theta}(x)$};

\draw[arrow] (data) -- (contrastive);
\draw[arrow] (contrastive) -- (embedding);


\node[method, below=10mm of data, xshift=3mm] (A)
{\textbf{A}\\Embedding LR};

\node[method, right=5mm of A] (B)
{\textbf{B}\\+ Cluster distances};

\node[method, right=5mm of B] (C)
{\textbf{C}\\+ High-FN regions};

\node[method, right=5mm of C] (D)
{\textbf{D}\\+ Dual-class structure};

\node[method, right=5mm of D] (E)
{\textbf{JDPI}\\Prototype inference};


\node[
    draw,
    rectangle,
    fit=(A)(B)(C)(D)(E),
    inner sep=6pt,
    label={[font=\bfseries]above:Detection in representation space}
] (detectorblock) {};

\draw[arrow] (embedding.south) |- ([xshift=15mm]detectorblock.north);

\draw[arrow] (A) -- (B);
\draw[arrow] (B) -- (C);
\draw[arrow] (C) -- (D);


\node[box, below=8 mm of C] (output)
{Harmful / Benign};

\draw[arrow] (A.south) |- (output.west);
\draw[arrow] (B.south) -- ++(0,-4mm) -| (output.west);
\draw[arrow] (C) -- (output);
\draw[arrow] (D.south) -- ++(0,-4mm) -| (output.east);
\draw[arrow] (E.south) |- (output.east);

\end{tikzpicture}

\caption{
Overview of the proposed framework.
Contrastive fine-tuning first learns safety-aware representations that increase the separation between harmful and benign inputs.
Method A serves as the embedding-based classifier baseline. Methods B--D progressively incorporate richer structural information, whereas JDPI performs direct prototype-based inference without relying on an additional downstream classifier.
}
\label{fig:pipeline}
\end{figure*}
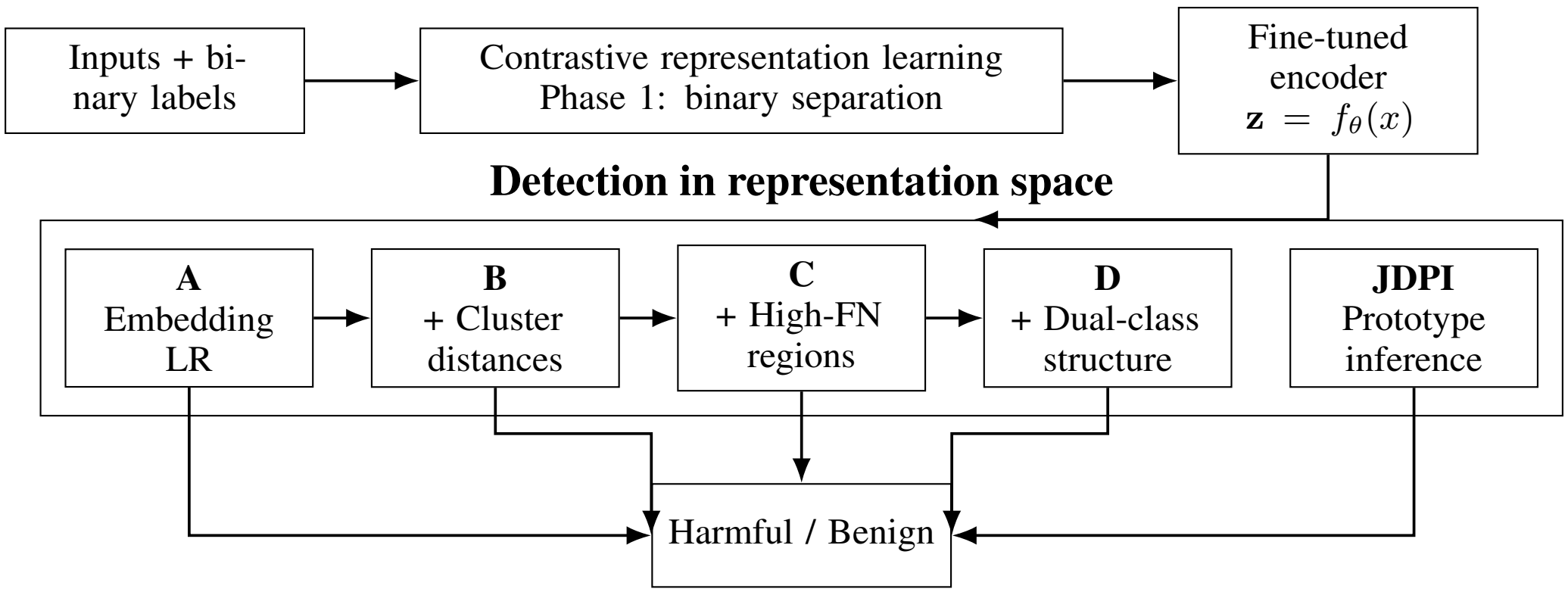

The proposed framework consists of two main stages. First, contrastive representation learning is employed to increase the separation between harmful and benign inputs in the embedding space, thereby inducing a safety-aware geometric structure. Second, the resulting representation space is exploited for downstream detection through a series of increasingly structure-aware approaches, ranging from conventional classifier-based methods to direct prototype-based inference. Figure~\ref{fig:pipeline} provides an overview of the framework.

\subsection{Problem formulation}

Given a labeled dataset $\mathcal{D}=\{(x_i,y_i)\}_{i=1}^{N}$ with $y_i\in\{0,1\}$, where $1$ denotes harmful and $0$ denotes benign, we encode each input as
$\mathbf{z}_i=f_{\theta}(x_i)\in\mathbb{R}^{d}$ \cite{song2020mpnetmaskedpermutedpretraining, reimers2019sentencebertsentenceembeddingsusing}.

\subsection{Contrastive Representation Learning}

We fine-tune the sentence encoder through contrastive
learning to improve harmful--benign separation \cite{khosla2021supervisedcontrastivelearning, gao2022simcsesimplecontrastivelearning, tunstall2022efficientfewshotlearningprompts, li2020sentenceembeddingspretrainedlanguage}. For each training pair $p=(x_p^{(1)},x_p^{(2)})$, we compute the cosine similarity
$s_p=\operatorname{sim}(f_\theta(x_p^{(1)}),f_\theta(x_p^{(2)}))$.
In the first phase, harmful--harmful and benign--benign pairs are treated as positive pairs and assigned label $1$, while harmful--benign pairs are treated as negative pairs and assigned label $0$ \cite{khosla2021supervisedcontrastivelearning, tunstall2022efficientfewshotlearningprompts}. The encoder is optimized using CoSENTLoss \cite{10531646}:

\begin{equation}
\mathcal{L}_{\mathrm{CoSENT}}
=
\log\!\left(
1+
\sum_{p\in\mathcal{P}^{+}}
\sum_{n\in\mathcal{P}^{-}}
\exp\!\left[\lambda\left(s_n-s_p\right)\right]
\right),
\label{eq:cosent}
\end{equation}
where $\mathcal{P}^{+}$ and $\mathcal{P}^{-}$ denote
the sets of positive and negative training pairs in a mini-batch, respectively, and $\lambda$ controls the scale of pairwise similarity differences.

We also evaluate an iterative refinement stage for harmful
subclusters, described in Appendix~\ref{app:refinement}.

\subsection{Structure-Aware detection method}

Methods A--D examine whether clustering-derived structural features improve supervised detection beyond using the adapted embeddings alone.

Method A is the embedding-based baseline \cite{galinkin2024improvedlargelanguagemodel}.
It predicts harmfulness directly from the embedding:
$p_A(y=1\mid\mathbf{z})
=\sigma(\mathbf{w}^{\top}\mathbf{z}+b)$.
The decision threshold $\tau_A$ is selected on the
validation split.

Method B augments the embedding with distances to
harmful cluster centroids. We apply $K$-means to harmful
examples in the structure split
\cite{10.1109/TIT.1982.1056489},
obtaining centroids
$\{\boldsymbol{\mu}^{h}_k\}_{k=1}^{K_h}$.
The augmented representation is
\begin{equation}
\begin{aligned}
\mathbf{d}^{h}(\mathbf{z})
&=\bigl[
\|\mathbf{z}-\boldsymbol{\mu}^{h}_1\|_2,\ldots,
\|\mathbf{z}-\boldsymbol{\mu}^{h}_{K_h}\|_2
\bigr],\\
\widetilde{\mathbf{z}}^{(B)}
&=[\mathbf{z};\mathbf{d}^{h}(\mathbf{z})].
\end{aligned}
\label{eq:cluster_features}
\end{equation}
This comparison tests whether harmful-cluster geometry
adds predictive information to the embedding.

Method C adds features describing regions with elevated
false-negative rates. For each harmful cluster $k$,
$r_k^{\mathrm{FN}}$ is estimated from Method B predictions
on the structure split. Clusters whose false-negative
rate exceeds the overall harmful-class rate and whose
size is at least $n_{\min}$ form the high-FN set
$\mathcal{H}_{\mathrm{FN}}$.
If no cluster satisfies the high-FN criterion, we use
the two clusters with the highest false-negative rates
as $\mathcal{H}_{\mathrm{FN}}$.
The remaining harmful clusters form
$\mathcal{L}_{\mathrm{FN}}$.
If this set is empty, $\mathcal{H}_{\mathrm{FN}}$ is
also used as the reference set for the low-FN features.

For an input representation, $d_H$ and $d_L$ denote its
minimum distances to these two sets, while $s_H$ and
$s_L$ are their aggregated distance-based soft-membership
scores. The indicator $I_H(\mathbf{z})$ records whether
the nearest harmful centroid belongs to
$\mathcal{H}_{\mathrm{FN}}$.
These quantities form
\begin{equation}
\begin{aligned}
\mathbf{r}_{\mathrm{FN}}(\mathbf{z})
=\bigl[&
d_H,d_L,d_H-d_L,\\
&s_H,s_L,s_H-s_L,I_H(\mathbf{z})
\bigr].
\end{aligned}
\label{eq:fn_features}
\end{equation}
Method C uses
$\widetilde{\mathbf{z}}^{(C)}
=[\widetilde{\mathbf{z}}^{(B)};
\mathbf{r}_{\mathrm{FN}}(\mathbf{z})]$.

Method D incorporates benign structure and
cross-class relationships through three feature groups.
The first, $\mathbf{r}_{HB}(\mathbf{z})$, contains
distance and soft-membership statistics for high-FN
harmful regions, all harmful clusters, and benign
clusters, together with their differences and ratios.
For the second group,
$\mathbf{r}_{\mathrm{mirror}}(\mathbf{z})$,
each high-FN harmful centroid is paired with its nearest benign centroid in Euclidean distance. Features include the input's distances to both centroids, their difference and ratio, and the difference between distance-based similarity scores computed using exponential decay.

The third group, $\mathbf{q}(\mathbf{z})$, contains
nine features derived from Methods A--C: each method's
harmful probability $p$, distance from the decision
midpoint $|p-0.5|$, and uncertainty score
$1-2|p-0.5|$.
Training features use five-fold out-of-fold predictions;
validation and test features use the corresponding
models fitted on the training split.
The final representation is
\begin{equation}
\begin{aligned}
\widetilde{\mathbf{z}}^{(D)}
=\bigl[&
\widetilde{\mathbf{z}}^{(C)};
\mathbf{r}_{HB}(\mathbf{z});\\
&\mathbf{r}_{\mathrm{mirror}}(\mathbf{z});
\mathbf{q}(\mathbf{z})
\bigr].
\end{aligned}
\label{eq:method_d}
\end{equation}

\subsection{Jailbreak Detection with Prototype Inference}

We refer to prototype-based inference (Method~E) as JDPI. JDPI examines whether the geometry of the same representation space can instead support detection directly. It makes the harmful--benign decision from relative distances to harmful and benign prototypes, without training an additional downstream classifier.

Following prototype-based metric learning \cite{snell2017prototypicalnetworksfewshotlearning},
we independently cluster harmful and benign examples in the structure split to obtain
$\mathcal{P}_h
=\{\boldsymbol{\mu}^{h}_k\}_{k=1}^{K_h}$
and
$\mathcal{P}_b
=\{\boldsymbol{\mu}^{b}_j\}_{j=1}^{K_b}$.
Both sets are constructed using binary safety labels.

For an input representation $\mathbf{z}$, its distances to the two prototype sets are
\begin{equation}
\begin{aligned}
d_h(\mathbf{z})
&=\min_k d_{\cos}
(\mathbf{z},\boldsymbol{\mu}^{h}_k),\\
d_b(\mathbf{z})
&=\min_j d_{\cos}
(\mathbf{z},\boldsymbol{\mu}^{b}_j),
\end{aligned}
\label{eq:prototype_distances}
\end{equation}
where $d_{\cos}$ denotes cosine distance.
A two-class softmax converts these distances into a harmfulness score:
\begin{equation}
p_E(y=1\mid\mathbf{z})
=
\frac{
\exp\!\left(-d_h(\mathbf{z})/T\right)
}{
\exp\!\left(-d_h(\mathbf{z})/T\right)
+
\exp\!\left(-d_b(\mathbf{z})/T\right)
}.
\label{eq:prototype_inference}
\end{equation}

The temperature $T$ controls the sharpness of the score. An input is classified as harmful when $p_E(y=1\mid\mathbf{z})\geq\tau_E$. Both $T$ and $\tau_E$ are selected on the validation split.

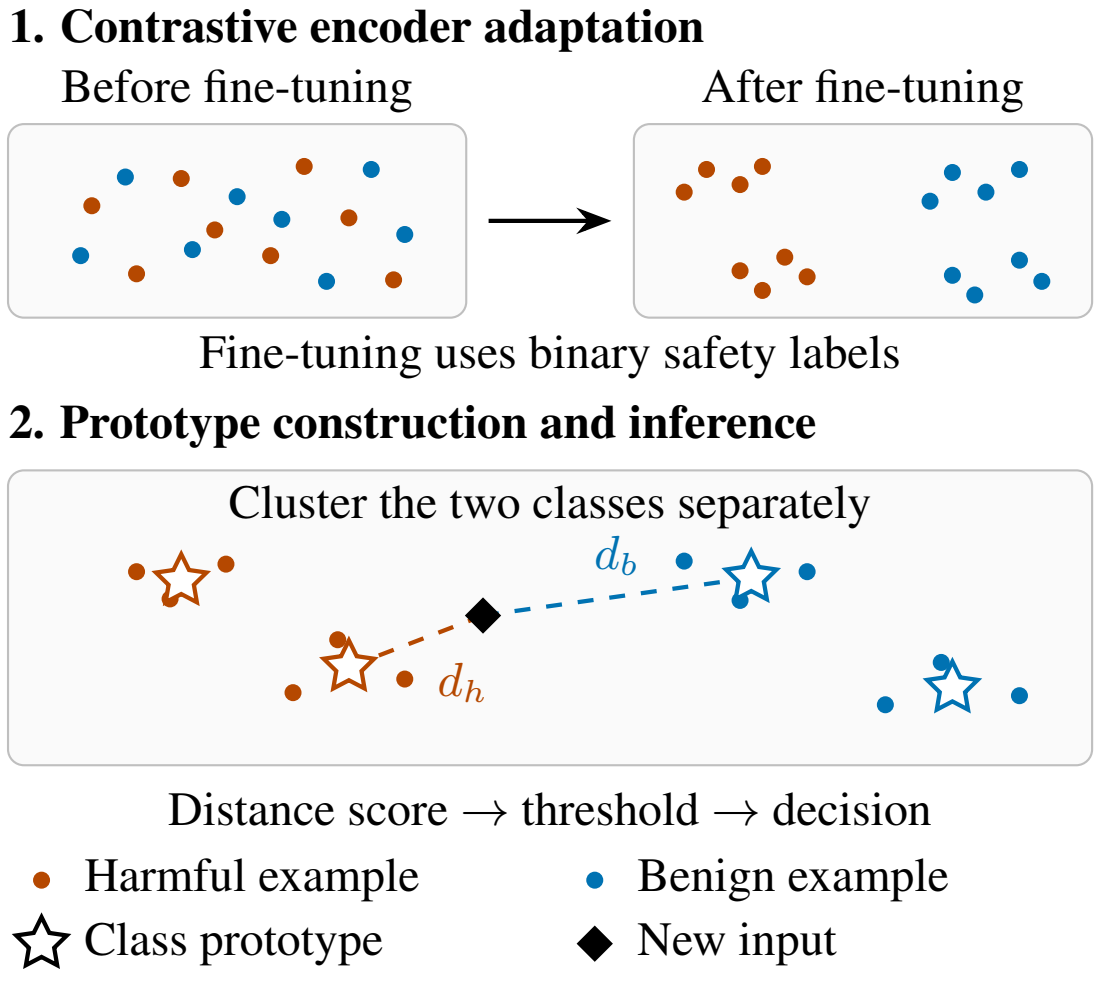
\begin{figure}[t]
\centering
\definecolor{jdpiharm}{HTML}{B54800}
\definecolor{jdpibenign}{HTML}{0072B2}

\begin{tikzpicture}[
    x=.96\columnwidth, y=1cm,
    font=\small,
    >=Stealth,
    panel/.style={draw=black!25, fill=black!2, rounded corners=3pt},
    proto/.style={
        star, star points=5, star point ratio=2,
        minimum size=10pt, inner sep=0pt, fill=white, line width=.8pt
    },
    query/.style={
        diamond, fill=black, minimum size=7pt, inner sep=0pt
    }
]

\node[anchor=west, inner sep=0pt] at (.015,6.40)
    {\textbf{1. Contrastive encoder adaptation}};

\node at (.22,6.00) {Before fine-tuning};
\node at (.78,6.00) {After fine-tuning};

\draw[panel] (.015,4.50) rectangle (.425,5.78);
\draw[panel] (.575,4.50) rectangle (.985,5.78);
\draw[->, thick] (.445,5.14) -- (.555,5.14);

\foreach \x/\y in {
    .09/5.24,.13/4.79,.17/5.42,.20/5.08,
    .25/4.91,.28/5.50,.32/5.16,.36/4.75
} {\fill[jdpiharm] (\x,\y) circle[radius=1.6pt];}
\foreach \x/\y in {
    .08/4.91,.12/5.43,.18/4.95,.22/5.30,
    .26/5.15,.30/4.74,.34/5.48,.37/5.05
} {\fill[jdpibenign] (\x,\y) circle[radius=1.6pt];}

\foreach \x/\y in {
    .62/5.33,.64/5.48,.67/5.38,.69/5.50,
    .67/4.81,.69/4.68,.71/4.90,.73/4.77
} {\fill[jdpiharm] (\x,\y) circle[radius=1.6pt];}
\foreach \x/\y in {
    .84/5.27,.86/5.46,.89/5.33,.92/5.48,
    .86/4.78,.88/4.65,.92/4.88,.94/4.74
} {\fill[jdpibenign] (\x,\y) circle[radius=1.6pt];}

\node at (.50,4.24) {Fine-tuning uses binary safety labels};

\node[anchor=west, inner sep=0pt] at (.015,3.78)
    {\textbf{2. Prototype construction and inference}};

\draw[panel] (.015,1.54) rectangle (.985,3.49);
\node at (.50,3.25) {Cluster the two classes separately};

\foreach \x/\y in {
    .13/2.82,.16/2.64,.21/2.87,
    .27/2.02,.31/2.37,.37/2.11
} {\fill[jdpiharm] (\x,\y) circle[radius=1.6pt];}
\foreach \x/\y in {
    .62/2.89,.67/2.63,.73/2.82,
    .80/1.94,.85/2.22,.92/2.00
} {\fill[jdpibenign] (\x,\y) circle[radius=1.6pt];}

\coordinate (jdpi-h) at (.32,2.19);
\coordinate (jdpi-b) at (.68,2.78);
\coordinate (jdpi-q) at (.44,2.53);

\draw[jdpiharm, dashed, thick]
    (jdpi-q) -- (jdpi-h) node[midway, below right] {$d_h$};
\draw[jdpibenign, dashed, thick]
    (jdpi-q) -- (jdpi-b) node[midway, above] {$d_b$};

\node[proto, draw=jdpiharm] at (.17,2.76) {};
\node[proto, draw=jdpiharm] at (jdpi-h) {};
\node[proto, draw=jdpibenign] at (jdpi-b) {};
\node[proto, draw=jdpibenign] at (.86,2.05) {};
\node[query] at (jdpi-q) {};

\node at (.50,1.25)
    {Distance score $\to$ threshold $\to$ decision};

\fill[jdpiharm] (.045,.78) circle[radius=1.6pt];
\node[anchor=west] at (.065,.78) {Harmful example};
\fill[jdpibenign] (.54,.78) circle[radius=1.6pt];
\node[anchor=west] at (.56,.78) {Benign example};

\node[proto, draw=black] at (.045,.36) {};
\node[anchor=west] at (.065,.36) {Class prototype};
\node[query] at (.54,.36) {};
\node[anchor=west] at (.56,.36) {New input};

\end{tikzpicture}
\caption{Overview of JDPI. Contrastive fine-tuning improves
harmful--benign separation. JDPI constructs prototypes from
each class in the structure split and scores new inputs using
the cosine distances $d_h$ and $d_b$ to the nearest prototypes
of the two classes. The point clouds are schematic.}
\label{fig:jdpi_geometry}
\end{figure}

Figure~\ref{fig:jdpi_geometry} illustrates how contrastive fine-tuning shapes the embedding space for direct prototype inference.

\section{Experiments}
\subsection{Datasets and Splits}
We evaluate safety detection on four benchmarks using individual prompts, question--answer pairs, and multi-turn conversations. The evaluation includes adversarial prompts from WildGuardMix and WildJailbreak and multi-turn jailbreak attempts from TomGibbs. Class proportions below refer to our experimental test splits. We abbreviate WildGuardMix, WildJailbreak, BeaverTailsQA, and TomGibbs as WGM, WJB, BTQA, and TG, respectively. We derive the binary safety labels from the released annotations or categories of each benchmark and match the unit of evaluation to the unit those labels describe.

\paragraph{WildGuardMix.}
WildGuardMix \cite{wildguard2024} contains vanilla and adversarial prompts with harmful and benign labels.
We use the prompt text and its prompt-harmfulness annotation. The test split contains 53.10\% harmful and 46.90\% benign prompts.

\paragraph{WildJailbreak.}
WildJailbreak \cite{jiang2024wildteamingscaleinthewildjailbreaks} contains vanilla and adversarial versions of harmful and benign requests, including benign prompts that resemble attacks but have no harmful intent.
Our classifier-training split combines 30,000 vanilla and 6,000 adversarial examples; the structure, validation, and test splits contain adversarial examples. The test split contains 51.77\% harmful and 48.23\% benign prompts.

\paragraph{BeaverTailsQA.}
BeaverTails \cite{ji2023beavertailsimprovedsafetyalignment} provides a safety label for each question--answer pair. The label depends on the answer in the context of the question, so the same question can receive different labels when paired with different answers. We therefore concatenate the user question and the assistant response as a single input. Unsafe pairs are mapped to the harmful class and safe pairs to the benign class, accounting for 55.42\% and 44.58\% of the test split, respectively.

\paragraph{TomGibbs.}
We use the input-only Multi-Turn Jailbreak Attack Datasets of \citet{gibbs2024emergingvulnerabilitiesfrontiermodels}. Each example provides a complete single-turn prompt and a corresponding sequence of user turns. The attacks use word-substitution ciphers and distribute their instructions across turns.

The dataset includes harmful, completely benign, and semi-benign examples. Semi-benign examples contain harmful or toxic vocabulary but pursue benign goals; both benign subsets are assigned label 0. These three groups account for 63.39\%, 18.75\%, and 17.86\% of our test split, respectively. For the separate static evaluation, we use the corresponding complete prompts from the same held-out examples.

\paragraph{Data splits.}

We divide each dataset into four splits. The structure split is used to fine-tune the encoder and construct clusters and class prototypes. The training split supplies the labeled examples for fitting the downstream classifiers. We use the validation split to select detector settings, including decision thresholds. The test split is held out throughout model development and used only for final evaluation.

For WildGuardMix and BeaverTailsQA, we pool the released training and test partitions, sample 60,000 examples stratified by the binary label, and construct new experimental splits. For TomGibbs, we assign entire goal groups to splits and downsample the training and structure splits to balance the binary labels. Table~\ref{tab:dataset_splits} summarizes the split sizes.

\begin{table}[t]
\centering
\begingroup
\small
\setlength{\tabcolsep}{2.5pt}
\begin{tabular}{@{}lrrrr@{}}
\toprule
Dataset & Train & Struct. & Val. & Test \\
\midrule
WildGuardMix  & 36,000 & 7,919 & 7,927 & 8,154 \\
WildJailbreak & 36,000 & 6,000 & 5,994 & 12,006 \\
BeaverTailsQA & 36,000 & 7,919 & 7,927 & 8,154 \\
TomGibbs      & 2,880  & 624   & 864   & 896 \\
\bottomrule
\end{tabular}
\endgroup
\caption{Experimental split sizes. Struct.\ denotes the
structure split used for representation adaptation and
prototype construction; Val.\ denotes validation.}
\label{tab:dataset_splits}
\end{table}

\subsection{Baselines and Evaluation Metrics}
We compare JDPI with an embedding-based classifier (Method~A)
using the same encoder. Methods B--D provide additional comparisons using structure-derived features. Llama Guard 3-8B serves as an external baseline without task-specific fine-tuning. We report accuracy and ROC-AUC, supplemented by F1, false-negative rate, and false-positive rate. We also report silhouette scores to characterize harmful--benign separation and clustering structure within the harmful class.

By avoiding the separate classifier-training split, JDPI uses 65.9--75.0\% fewer labeled examples for model development than classifier~A across the four benchmarks (Table~\ref{tab:label_budget}). In cross-dataset evaluation, the encoder, classifier or prototypes, and selected parameters are obtained from the source dataset and applied to the target test split without target-dataset adaptation.

For fixed multi-turn replay on TomGibbs, we compare scoring the current user turn alone with scoring the accumulated user history up to and including that turn. Each method uses the detector and threshold selected in the complete-prompt setting throughout replay. Following the any-turn blocking criterion used by \citet{gibbs2024emergingvulnerabilitiesfrontiermodels}, we count a conversation as blocked if the detector flags the input at any evaluated user turn.
\subsection{Implementation Details}

We use a random seed of 42. The sentence encoder is \texttt{all-mpnet-base-v2} from Sentence Transformers, with an encoding batch size of 64 for downstream evaluation. Contrastive fine-tuning uses CoSENTLoss. Phase~1 uses five epochs and 200 warm-up steps. Phase~2 runs for at most five iterations, with two epochs and 50 warm-up steps per iteration. Appendix~\ref{app:implementation} reports the encoder training configurations and contrastive pair counts.

For clustering, $K$ is selected from $\{4,\ldots,12\}$ using the cosine-distance silhouette score. K-means uses 10 initializations. The maximum iteration counts for classifiers A--D are 3,000, 4,000, 5,000, and 6,000, respectively. Methods A--C use balanced class weights. Method D constructs its training prediction features through five-fold out-of-fold prediction.

For Methods A--C and JDPI, decision thresholds are selected on the validation split from $\{0.10,0.15,\ldots,0.90\}$. JDPI additionally selects its temperature from $\{0.1,0.3,0.5,0.7,1.0,1.5,2.0\}$. Threshold and temperature selection for JDPI maximizes validation accuracy. Methods B--D use Euclidean cluster-distance features, while JDPI uses cosine distances to prototypes.

For the external baseline, we use \texttt{Llama-Guard-3-8B} with greedy decoding and at most 10 generated tokens. Outputs beginning with \texttt{unsafe} are mapped to the harmful class; remaining outputs are mapped to the benign class. Continuous unsafe scores are used for ROC-AUC calculation, separately from these binary decisions.

\section{Results and Analysis}

\subsection{Main Detection Results}
Table~\ref{tab:main_results} compares direct embedding classification (A), classifiers augmented with structural features (B--D), and JDPI.
Methods B--D provide modest and dataset-dependent improvements over A, with no single structural variant consistently performing best. These results suggest that much of the discriminative information is already accessible to the classifier through the adapted embeddings. JDPI examines a complementary question: whether this information can support detection through prototype distances without training an additional downstream classifier.

\begin{table}[t]
\centering
\begingroup
\small
\setlength{\tabcolsep}{3pt}
\begin{tabular}{@{}lrrrrr@{}}
\toprule
Metric & A & B & C & D & JDPI \\
\midrule
\multicolumn{6}{@{}l}{\textbf{WildGuardMix}} \\
Acc. & 94.43 & 94.55 & 94.49 & \textbf{94.69} & 94.13 \\
AUC  & .9874 & .9876 & .9877 & \textbf{.9878} & .9860 \\
\midrule
\multicolumn{6}{@{}l}{\textbf{WildJailbreak}} \\
Acc. & \textbf{95.54} & 95.49 & 95.52 & 95.36 & 95.24 \\
AUC  & .9906 & \textbf{.9909} & .9907 & .9908 & .9900 \\
\midrule
\multicolumn{6}{@{}l}{\textbf{BeaverTails  QA}} \\
Acc. & \textbf{82.24} & 82.02 & 82.06 & 82.23 & 80.48 \\
AUC  & .8938 & .8943 & .8949 & \textbf{.8954} & .8772 \\
\bottomrule
\end{tabular}
\endgroup
\caption{In-domain results of Methods A--E using the same fine-tuned encoder within each dataset. Accuracy is reported as a percentage. Bold indicates the highest value in each row.}
\label{tab:main_results}
\end{table}

JDPI closely matches the baseline on WildGuardMix and WildJailbreak.
Compared with classifier~A, JDPI's accuracy is 0.30 percentage points lower on both WildGuardMix and WildJailbreak. Its ROC-AUC is lower by 0.0014 on WildGuardMix and 0.0006 on WildJailbreak. After contrastive fine-tuning improves the separation between harmful and benign examples in the embedding space, JDPI uses prototype distances to achieve performance close to that of classifier~A on both benchmarks.

On BeaverTailsQA, classifier~A exceeds JDPI by 1.76 percentage points in accuracy and by 0.0166 in ROC-AUC. We report the TomGibbs conversation-level results separately under the any-turn rule. On TomGibbs, classifier~A achieves higher accuracy in the complete-prompt evaluation reported in the appendix. In the any-turn replay, JDPI achieves higher conversation-level accuracy when both methods use accumulated user history.

The central finding is that an additional downstream classifier is not always necessary to obtain competitive detection performance after contrastive adaptation. The remaining gaps also identify settings in which a learned decision function captures useful information beyond nearest-prototype distances. Figure~\ref{fig:ae_disagreement} in the appendix shows which test examples JDPI and classifier~A classify correctly.

\subsection{Fixed Multi-Turn Replay Evaluation} 

We also evaluate detection as multi-turn conversations unfold. In TomGibbs, we compare scoring each user turn in isolation with scoring it using the accumulated user history (Table~\ref{tab:multiturn}).

\begin{table}[t]
\centering
\begingroup
\small
\setlength{\tabcolsep}{4pt}
\begin{tabular}{@{}llrrr@{}}
\toprule
Input & Method & Acc.$\uparrow$ & FNR$\downarrow$ & FPR$\downarrow$ \\
\midrule
Current & A & 63.39 & 0.00 & 100.00 \\
        & JDPI & 77.23 & 6.34 & 51.22 \\
\midrule
History & A & 63.39 & 0.00 & 100.00 \\
        & JDPI & 89.29 & 4.23 & 21.95 \\
\bottomrule
\end{tabular}
\endgroup
\caption{Any-turn detection on TomGibbs (\%).}
\label{tab:multiturn}
\end{table}

Under the any-turn rule, JDPI improves conversation-level accuracy over classifier~A by 13.8 percentage points with current-turn inputs and by 25.9 points with accumulated user history. With accumulated history, JDPI lowers the false-positive rate by 78.05 percentage points while raising the false-negative rate by 4.23 points. At its fixed threshold, classifier~A flags every conversation in both input settings, yielding zero FNR but 100\% FPR. The reduction in false alarms more than offsets the increase in missed harmful conversations.

For JDPI, using accumulated user history increases conversation-level accuracy by 12.06 percentage points and reduces FPR and FNR by 29.27 and 2.11 percentage points, respectively, compared with using only the current turn. Under this fixed replay protocol, accumulated context therefore improves both harmful-conversation detection and benign-conversation acceptance for JDPI.

Appendix~\ref{app:finetuning_effect} reports the effect of contrastive fine-tuning.

\subsection{Comparison with External Safety-Guard}

We compare JDPI with Llama Guard 3-8B. JDPI uses an encoder and prototypes adapted to each benchmark, whereas Llama Guard uses its released weights without target-dataset fine-tuning. Both methods are evaluated on the same test splits, with ROC-AUC computed from continuous scores.

\begin{table}[t]
\centering
\begingroup
\small
\setlength{\tabcolsep}{3pt}
\begin{tabular}{@{}lrrrr@{}}
\toprule
& \multicolumn{2}{c}{Accuracy (\%)}
& \multicolumn{2}{c}{ROC-AUC} \\
\cmidrule(lr){2-3}
\cmidrule(lr){4-5}
Dataset & LG3 & JDPI & LG3 & JDPI \\
\midrule
WildGuardMix
& 84.04 & \textbf{94.13}
& 0.6781 & \textbf{0.9860} \\
WildJailbreak
& 71.24 & \textbf{95.24}
& 0.7284 & \textbf{0.9900} \\
BeaverTailsQA
& 71.08 & \textbf{80.48}
& 0.7043 & \textbf{0.8772} \\
TomGibbs
& 77.90 & \textbf{89.73}
& 0.6540 & \textbf{0.9637} \\
\bottomrule
\end{tabular}
\endgroup
\caption{Comparison with Llama Guard 3-8B (LG3).}
\label{tab:external_llamaguard}
\end{table}

As shown in Table~\ref{tab:external_llamaguard},
JDPI achieves higher accuracy and ROC-AUC on all four benchmarks. Accuracy gains range from 9.40 to 24.00 percentage points, while ROC-AUC gains range from 0.1729 to 0.3097. The largest accuracy gain occurs on WildJailbreak, where accuracy increases from 71.24\% to 95.24\%.

This comparison demonstrates the performance obtainable with dataset-adapted representations and direct prototype inference relative to the released guard. It complements the A/JDPI comparison, which holds the encoder fixed while comparing classifier-based and prototype-based decisions.

\subsection{Cross-Dataset Generalization}

\begin{figure}[t]
\centering
\includegraphics[width=\columnwidth]
{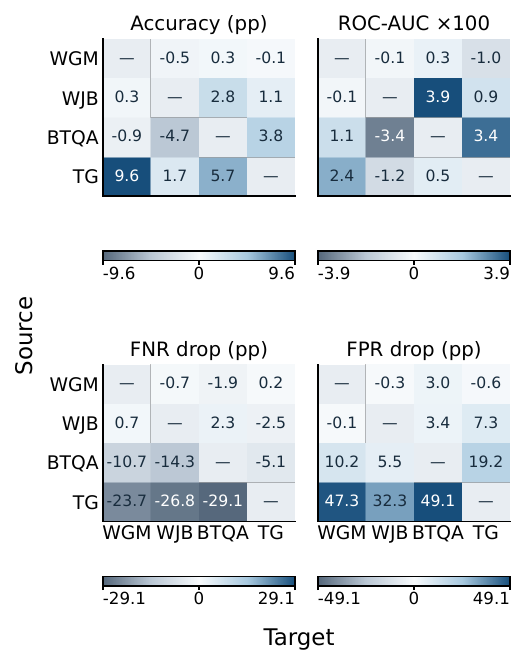}
\caption{Cross-dataset differences between Methods A and JDPI.
Accuracy and ROC-AUC show JDPI minus A; FNR and FPR show A minus JDPI. Positive values favor JDPI in every panel.
Rows denote sources and columns denote targets. ROC-AUC differences are multiplied by 100.}
\label{fig:cross_dataset}
\end{figure}

We evaluate cross-dataset transfer to examine whether source-adapted detectors generalize to other datasets. Figure~\ref{fig:cross_dataset} compares Methods A and JDPI across all 12 directed transfers. Rows represent source datasets and columns represent target datasets. The sign convention is chosen so that positive values favor JDPI in every panel.

JDPI achieves higher accuracy in eight of the 12 transfer directions. Across all 12 directions, weighted equally, JDPI exceeds classifier~A by 1.59 percentage points in mean accuracy and by 0.0055 in mean ROC-AUC. These results show that prototype inference can retain competitive cross-dataset performance without training an additional downstream classifier.

The gains depend on the source--target pair. With TomGibbs as the source, JDPI exceeds classifier~A in accuracy by 9.59 percentage points on WildGuardMix and 5.74 points on BeaverTailsQA. JDPI reduces mean FPR from 47.31\% to 32.63\%, while mean FNR increases from 16.74\% to 26.05\%. Appendix~\ref{app:cross_dataset} reports the full results.

\section{Conclusion}
This paper investigates how the structure of learned representations can support jailbreak and harmful-input detection. We examine two related questions: whether explicit structural information improves a conventional embedding-based classifier, and whether the learned geometry can support detection without training an additional downstream classifier.

Contrastive fine-tuning improves the separation between harmful and benign examples in the embedding space, providing a clearer structure for constructing class prototypes. It also improves JDPI's detection accuracy across all four benchmarks. The clustering-derived features in Methods~B--D yield small, dataset-dependent gains in the in-domain experiments. This suggests that classifier~A already captures much of the signal needed for detection directly from the fine-tuned embeddings. On WildGuardMix, WildJailbreak, and BeaverTailsQA, JDPI achieves accuracy close to that of classifier~A while saving the labeled data and computation required to train a downstream classifier. With dataset-specific adaptation, JDPI exceeds the released Llama Guard 3-8B baseline in accuracy and ROC-AUC on all four benchmarks. Together, these results support representation adaptation as a basis for effective detection with a simple downstream decision rule.

Across the 12 directed transfers, JDPI achieves higher mean
accuracy than classifier~A, suggesting greater robustness to
dataset shift on this metric. In multi-turn replay on TomGibbs, retaining user history improves JDPI's conversation-level accuracy, which exceeds that of classifier~A under the same input setting. Together, these results show that contrastively adapted representations can support competitive prototype-based detection. JDPI uses binary safety labels for encoder adaptation, prototype construction, and validation. JDPI avoids the separate labeled split used to train classifier~A and shows greater robustness to dataset shift as measured by mean accuracy across the evaluated transfers.

\section*{Limitations}

Although JDPI removes the need for a trained downstream classifier, binary labels are still required for contrastive representation learning, class-specific prototype construction, and validation-based parameter selection. In our experimental setup, it avoids the additional labeled training split used by the classifier, but does not eliminate dependence on labeled data.

How larger fine-tuning datasets or greater encoder training budgets affect the relative performance of JDPI and the classifier baseline remains an open question. Although the decision threshold and temperature of JDPI are selected on the validation split, the search covers a finite set of candidate values. Sensitivity to these parameters, training duration, clustering configurations, and distance metrics has not been fully characterized. 

Cross-dataset experiments show that transfer performance varies across source--target pairs and involves different false-positive and false-negative trade-offs.

\section*{Acknowledgments}


\bibliography{custom}

@misc{ganguli2022redteaminglanguagemodels,
      title={Red Teaming Language Models to Reduce Harms: Methods, Scaling Behaviors, and Lessons Learned}, 
      author={Deep Ganguli and Liane Lovitt and Jackson Kernion and Amanda Askell and Yuntao Bai and Saurav Kadavath and Ben Mann and Ethan Perez and Nicholas Schiefer and Kamal Ndousse and Andy Jones and Sam Bowman and Anna Chen and Tom Conerly and Nova DasSarma and Dawn Drain and Nelson Elhage and Sheer El-Showk and Stanislav Fort and Zac Hatfield-Dodds and Tom Henighan and Danny Hernandez and Tristan Hume and Josh Jacobson and Scott Johnston and Shauna Kravec and Catherine Olsson and Sam Ringer and Eli Tran-Johnson and Dario Amodei and Tom Brown and Nicholas Joseph and Sam McCandlish and Chris Olah and Jared Kaplan and Jack Clark},
      year={2022},
      eprint={2209.07858},
      archivePrefix={arXiv},
      primaryClass={cs.CL},
      url={https://arxiv.org/abs/2209.07858}, 
}

@misc{sharma2025constitutionalclassifiersdefendinguniversal,
      title={Constitutional Classifiers: Defending against Universal Jailbreaks across Thousands of Hours of Red Teaming}, 
      author={Mrinank Sharma and Meg Tong and Jesse Mu and Jerry Wei and Jorrit Kruthoff and Scott Goodfriend and Euan Ong and Alwin Peng and Raj Agarwal and Cem Anil and Amanda Askell and Nathan Bailey and Joe Benton and Emma Bluemke and Samuel R. Bowman and Eric Christiansen and Hoagy Cunningham and Andy Dau and Anjali Gopal and Rob Gilson and Logan Graham and Logan Howard and Nimit Kalra and Taesung Lee and Kevin Lin and Peter Lofgren and Francesco Mosconi and Clare O'Hara and Catherine Olsson and Linda Petrini and Samir Rajani and Nikhil Saxena and Alex Silverstein and Tanya Singh and Theodore Sumers and Leonard Tang and Kevin K. Troy and Constantin Weisser and Ruiqi Zhong and Giulio Zhou and Jan Leike and Jared Kaplan and Ethan Perez},
      year={2025},
      eprint={2501.18837},
      archivePrefix={arXiv},
      primaryClass={cs.CL},
      url={https://arxiv.org/abs/2501.18837}, 
}

@misc{inan2023llamaguardllmbasedinputoutput,
      title={Llama Guard: LLM-based Input-Output Safeguard for Human-AI Conversations}, 
      author={Hakan Inan and Kartikeya Upasani and Jianfeng Chi and Rashi Rungta and Krithika Iyer and Yuning Mao and Michael Tontchev and Qing Hu and Brian Fuller and Davide Testuggine and Madian Khabsa},
      year={2023},
      eprint={2312.06674},
      archivePrefix={arXiv},
      primaryClass={cs.CL},
      url={https://arxiv.org/abs/2312.06674}, 
}

@misc{cui2025orbenchoverrefusalbenchmarklarge,
      title={OR-Bench: An Over-Refusal Benchmark for Large Language Models}, 
      author={Justin Cui and Wei-Lin Chiang and Ion Stoica and Cho-Jui Hsieh},
      year={2025},
      eprint={2405.20947},
      archivePrefix={arXiv},
      primaryClass={cs.CL},
      url={https://arxiv.org/abs/2405.20947}, 
}

@misc{wallace2024instructionhierarchytrainingllms,
      title={The Instruction Hierarchy: Training LLMs to Prioritize Privileged Instructions}, 
      author={Eric Wallace and Kai Xiao and Reimar Leike and Lilian Weng and Johannes Heidecke and Alex Beutel},
      year={2024},
      eprint={2404.13208},
      archivePrefix={arXiv},
      primaryClass={cs.CR},
      url={https://arxiv.org/abs/2404.13208}, 
}

@misc{guan2025deliberativealignmentreasoningenables,
      title={Deliberative Alignment: Reasoning Enables Safer Language Models}, 
      author={Melody Y. Guan and Manas Joglekar and Eric Wallace and Saachi Jain and Boaz Barak and Alec Helyar and Rachel Dias and Andrea Vallone and Hongyu Ren and Jason Wei and Hyung Won Chung and Sam Toyer and Johannes Heidecke and Alex Beutel and Amelia Glaese},
      year={2025},
      eprint={2412.16339},
      archivePrefix={arXiv},
      primaryClass={cs.CL},
      url={https://arxiv.org/abs/2412.16339}, 
}

@misc{sheshadri2025latentadversarialtrainingimproves,
      title={Latent Adversarial Training Improves Robustness to Persistent Harmful Behaviors in LLMs}, 
      author={Abhay Sheshadri and Aidan Ewart and Phillip Guo and Aengus Lynch and Cindy Wu and Vivek Hebbar and Henry Sleight and Asa Cooper Stickland and Ethan Perez and Dylan Hadfield-Menell and Stephen Casper},
      year={2025},
      eprint={2407.15549},
      archivePrefix={arXiv},
      primaryClass={cs.LG},
      url={https://arxiv.org/abs/2407.15549}, 
}

@misc{zou2024improvingalignmentrobustnesscircuit,
      title={Improving Alignment and Robustness with Circuit Breakers}, 
      author={Andy Zou and Long Phan and Justin Wang and Derek Duenas and Maxwell Lin and Maksym Andriushchenko and Rowan Wang and Zico Kolter and Matt Fredrikson and Dan Hendrycks},
      year={2024},
      eprint={2406.04313},
      archivePrefix={arXiv},
      primaryClass={cs.LG},
      url={https://arxiv.org/abs/2406.04313}, 
}

@misc{robey2024smoothllmdefendinglargelanguage,
      title={SmoothLLM: Defending Large Language Models Against Jailbreaking Attacks}, 
      author={Alexander Robey and Eric Wong and Hamed Hassani and George J. Pappas},
      year={2024},
      eprint={2310.03684},
      archivePrefix={arXiv},
      primaryClass={cs.LG},
      url={https://arxiv.org/abs/2310.03684}, 
}

@misc{jain2023baselinedefensesadversarialattacks,
      title={Baseline Defenses for Adversarial Attacks Against Aligned Language Models}, 
      author={Neel Jain and Avi Schwarzschild and Yuxin Wen and Gowthami Somepalli and John Kirchenbauer and Ping-yeh Chiang and Micah Goldblum and Aniruddha Saha and Jonas Geiping and Tom Goldstein},
      year={2023},
      eprint={2309.00614},
      archivePrefix={arXiv},
      primaryClass={cs.LG},
      url={https://arxiv.org/abs/2309.00614}, 
}

@misc{shen2024donowcharacterizingevaluating,
      title={"Do Anything Now": Characterizing and Evaluating In-The-Wild Jailbreak Prompts on Large Language Models}, 
      author={Xinyue Shen and Zeyuan Chen and Michael Backes and Yun Shen and Yang Zhang},
      year={2024},
      eprint={2308.03825},
      archivePrefix={arXiv},
      primaryClass={cs.CR},
      url={https://arxiv.org/abs/2308.03825}, 
}

@misc{wei2023jailbrokendoesllmsafety,
      title={Jailbroken: How Does LLM Safety Training Fail?}, 
      author={Alexander Wei and Nika Haghtalab and Jacob Steinhardt},
      year={2023},
      eprint={2307.02483},
      archivePrefix={arXiv},
      primaryClass={cs.LG},
      url={https://arxiv.org/abs/2307.02483}, 
}

@misc{perez2022redteaminglanguagemodels,
      title={Red Teaming Language Models with Language Models}, 
      author={Ethan Perez and Saffron Huang and Francis Song and Trevor Cai and Roman Ring and John Aslanides and Amelia Glaese and Nat McAleese and Geoffrey Irving},
      year={2022},
      eprint={2202.03286},
      archivePrefix={arXiv},
      primaryClass={cs.CL},
      url={https://arxiv.org/abs/2202.03286}, 
}

@misc{souly2024strongrejectjailbreaks,
      title={A StrongREJECT for Empty Jailbreaks}, 
      author={Alexandra Souly and Qingyuan Lu and Dillon Bowen and Tu Trinh and Elvis Hsieh and Sana Pandey and Pieter Abbeel and Justin Svegliato and Scott Emmons and Olivia Watkins and Sam Toyer},
      year={2024},
      eprint={2402.10260},
      archivePrefix={arXiv},
      primaryClass={cs.LG},
      url={https://arxiv.org/abs/2402.10260}, 
}

@misc{chao2024jailbreakbenchopenrobustnessbenchmark,
      title={JailbreakBench: An Open Robustness Benchmark for Jailbreaking Large Language Models}, 
      author={Patrick Chao and Edoardo Debenedetti and Alexander Robey and Maksym Andriushchenko and Francesco Croce and Vikash Sehwag and Edgar Dobriban and Nicolas Flammarion and George J. Pappas and Florian Tramer and Hamed Hassani and Eric Wong},
      year={2024},
      eprint={2404.01318},
      archivePrefix={arXiv},
      primaryClass={cs.CR},
      url={https://arxiv.org/abs/2404.01318}, 
}

@misc{mazeika2024harmbenchstandardizedevaluationframework,
      title={HarmBench: A Standardized Evaluation Framework for Automated Red Teaming and Robust Refusal}, 
      author={Mantas Mazeika and Long Phan and Xuwang Yin and Andy Zou and Zifan Wang and Norman Mu and Elham Sakhaee and Nathaniel Li and Steven Basart and Bo Li and David Forsyth and Dan Hendrycks},
      year={2024},
      eprint={2402.04249},
      archivePrefix={arXiv},
      primaryClass={cs.LG},
      url={https://arxiv.org/abs/2402.04249}, 
}

@misc{zou2023universaltransferableadversarialattacks,
      title={Universal and Transferable Adversarial Attacks on Aligned Language Models}, 
      author={Andy Zou and Zifan Wang and Nicholas Carlini and Milad Nasr and J. Zico Kolter and Matt Fredrikson},
      year={2023},
      eprint={2307.15043},
      archivePrefix={arXiv},
      primaryClass={cs.CL},
      url={https://arxiv.org/abs/2307.15043}, 
}

@misc{jiang2024wildteamingscaleinthewildjailbreaks,
      title={WildTeaming at Scale: From In-the-Wild Jailbreaks to (Adversarially) Safer Language Models}, 
      author={Liwei Jiang and Kavel Rao and Seungju Han and Allyson Ettinger and Faeze Brahman and Sachin Kumar and Niloofar Mireshghallah and Ximing Lu and Maarten Sap and Yejin Choi and Nouha Dziri},
      year={2024},
      eprint={2406.18510},
      archivePrefix={arXiv},
      primaryClass={cs.CL},
      url={https://arxiv.org/abs/2406.18510}, 
}

@misc{jin2024jailbreakhuntervisualanalyticsapproach,
      title={JailbreakHunter: A Visual Analytics Approach for Jailbreak Prompts Discovery from Large-Scale Human-LLM Conversational Datasets}, 
      author={Zhihua Jin and Shiyi Liu and Haotian Li and Xun Zhao and Huamin Qu},
      year={2024},
      eprint={2407.03045},
      archivePrefix={arXiv},
      primaryClass={cs.HC},
      url={https://arxiv.org/abs/2407.03045}, 
}

@misc{ball2024understandingjailbreaksuccessstudy,
      title={Understanding Jailbreak Success: A Study of Latent Space Dynamics in Large Language Models}, 
      author={Sarah Ball and Frauke Kreuter and Nina Panickssery},
      year={2024},
      eprint={2406.09289},
      archivePrefix={arXiv},
      primaryClass={cs.CL},
      url={https://arxiv.org/abs/2406.09289}, 
}

@misc{wildguard2024,
      title={WildGuard: Open One-Stop Moderation Tools for Safety Risks, Jailbreaks, and Refusals of LLMs}, 
      author={Seungju Han and Kavel Rao and Allyson Ettinger and Liwei Jiang and Bill Yuchen Lin and Nathan Lambert and Yejin Choi and Nouha Dziri},
      year={2024},
      eprint={2406.18495},
      archivePrefix={arXiv},
      primaryClass={cs.CL},
      url={https://arxiv.org/abs/2406.18495}, 
}

@misc{ji2023beavertailsimprovedsafetyalignment,
      title={BeaverTails: Towards Improved Safety Alignment of LLM via a Human-Preference Dataset}, 
      author={Jiaming Ji and Mickel Liu and Juntao Dai and Xuehai Pan and Chi Zhang and Ce Bian and Chi Zhang and Ruiyang Sun and Yizhou Wang and Yaodong Yang},
      year={2023},
      eprint={2307.04657},
      archivePrefix={arXiv},
      primaryClass={cs.CL},
      url={https://arxiv.org/abs/2307.04657}, 
}

@misc{gao2022simcsesimplecontrastivelearning,
      title={SimCSE: Simple Contrastive Learning of Sentence Embeddings}, 
      author={Tianyu Gao and Xingcheng Yao and Danqi Chen},
      year={2022},
      eprint={2104.08821},
      archivePrefix={arXiv},
      primaryClass={cs.CL},
      url={https://arxiv.org/abs/2104.08821}, 
}

@misc{reimers2019sentencebertsentenceembeddingsusing,
      title={Sentence-BERT: Sentence Embeddings using Siamese BERT-Networks}, 
      author={Nils Reimers and Iryna Gurevych},
      year={2019},
      eprint={1908.10084},
      archivePrefix={arXiv},
      primaryClass={cs.CL},
      url={https://arxiv.org/abs/1908.10084}, 
}

@misc{chen2020simpleframeworkcontrastivelearning,
      title={A Simple Framework for Contrastive Learning of Visual Representations}, 
      author={Ting Chen and Simon Kornblith and Mohammad Norouzi and Geoffrey Hinton},
      year={2020},
      eprint={2002.05709},
      archivePrefix={arXiv},
      primaryClass={cs.LG},
      url={https://arxiv.org/abs/2002.05709}, 
}

@misc{khosla2021supervisedcontrastivelearning,
      title={Supervised Contrastive Learning}, 
      author={Prannay Khosla and Piotr Teterwak and Chen Wang and Aaron Sarna and Yonglong Tian and Phillip Isola and Aaron Maschinot and Ce Liu and Dilip Krishnan},
      year={2021},
      eprint={2004.11362},
      archivePrefix={arXiv},
      primaryClass={cs.LG},
      url={https://arxiv.org/abs/2004.11362}, 
}

@ARTICLE{10531646,
  author={Huang, Xiang and Peng, Hao and Zou, Dongcheng and Liu, Zhiwei and Li, Jianxin and Liu, Kay and Wu, Jia and Su, Jianlin and Yu, Philip S.},
  journal={IEEE/ACM Transactions on Audio, Speech, and Language Processing}, 
  title={CoSENT: Consistent Sentence Embedding via Similarity Ranking}, 
  year={2024},
  volume={32},
  number={},
  pages={2800-2813},
  doi={10.1109/TASLP.2024.3402087}}

@misc{devlin2019bertpretrainingdeepbidirectional,
      title={BERT: Pre-training of Deep Bidirectional Transformers for Language Understanding}, 
      author={Jacob Devlin and Ming-Wei Chang and Kenton Lee and Kristina Toutanova},
      year={2019},
      eprint={1810.04805},
      archivePrefix={arXiv},
      primaryClass={cs.CL},
      url={https://arxiv.org/abs/1810.04805}, 
}

@misc{tanwisuth2021prototypeorientedframeworkunsuperviseddomain,
      title={A Prototype-Oriented Framework for Unsupervised Domain Adaptation}, 
      author={Korawat Tanwisuth and Xinjie Fan and Huangjie Zheng and Shujian Zhang and Hao Zhang and Bo Chen and Mingyuan Zhou},
      year={2021},
      eprint={2110.12024},
      archivePrefix={arXiv},
      primaryClass={cs.LG},
      url={https://arxiv.org/abs/2110.12024}, 
}

@misc{snell2017prototypicalnetworksfewshotlearning,
      title={Prototypical Networks for Few-shot Learning}, 
      author={Jake Snell and Kevin Swersky and Richard S. Zemel},
      year={2017},
      eprint={1703.05175},
      archivePrefix={arXiv},
      primaryClass={cs.LG},
      url={https://arxiv.org/abs/1703.05175}, 
}

@misc{kadali2026jailbreakingleavestraceunderstanding,
      title={Jailbreaking Leaves a Trace: Understanding and Detecting Jailbreak Attacks from Internal Representations of Large Language Models}, 
      author={Sri Durga Sai Sowmya Kadali and Evangelos E. Papalexakis},
      year={2026},
      eprint={2602.11495},
      archivePrefix={arXiv},
      primaryClass={cs.CR},
      url={https://arxiv.org/abs/2602.11495}, 
}

@inproceedings{jiao-etal-2026-llm,
    title = "{LLM} Safety From Within: Detecting Harmful Content with Internal Representations",
    author = "Jiao, Difan  and
      Liu, Yilun  and
      Yuan, Ye  and
      Tang, Zhenwei  and
      Du, Linfeng  and
      Wu, Haolun  and
      Anderson, Ashton",
    editor = "Liakata, Maria  and
      Moreira, Viviane P.  and
      Zhang, Jiajun  and
      Jurgens, David",
    booktitle = "Proceedings of the 64th Annual Meeting of the {A}ssociation for {C}omputational {L}inguistics (Volume 1: Long Papers)",
    month = jul,
    year = "2026",
    address = "San Diego, California, United States",
    publisher = "Association for Computational Linguistics",
    url = "https://aclanthology.org/2026.acl-long.1844/",
    doi = "10.18653/v1/2026.acl-long.1844",
    pages = "39711--39727",
    ISBN = "979-8-89176-390-6"
}

@misc{zhang2025jbshielddefendinglargelanguage,
      title={JBShield: Defending Large Language Models from Jailbreak Attacks through Activated Concept Analysis and Manipulation}, 
      author={Shenyi Zhang and Yuchen Zhai and Keyan Guo and Hongxin Hu and Shengnan Guo and Zheng Fang and Lingchen Zhao and Chao Shen and Cong Wang and Qian Wang},
      year={2025},
      eprint={2502.07557},
      archivePrefix={arXiv},
      primaryClass={cs.CR},
      url={https://arxiv.org/abs/2502.07557}, 
}

@inproceedings{hua-etal-2026-rethinking,
    title = "Rethinking Jailbreak Detection of Large Vision Language Models with Representational Contrastive Scoring",
    author = "Hua, Peichun  and
      Li, Hao  and
      Shi, Shanghao  and
      Yu, Zhiyuan  and
      Zhang, Ning",
    editor = "Liakata, Maria  and
      Moreira, Viviane P.  and
      Zhang, Jiajun  and
      Jurgens, David",
    booktitle = "Proceedings of the 64th Annual Meeting of the {A}ssociation for {C}omputational {L}inguistics (Volume 1: Long Papers)",
    month = jul,
    year = "2026",
    address = "San Diego, California, United States",
    publisher = "Association for Computational Linguistics",
    url = "https://aclanthology.org/2026.acl-long.992/",
    doi = "10.18653/v1/2026.acl-long.992",
    pages = "21748--21785",
    ISBN = "979-8-89176-390-6"
}

@misc{song2020mpnetmaskedpermutedpretraining,
      title={MPNet: Masked and Permuted Pre-training for Language Understanding}, 
      author={Kaitao Song and Xu Tan and Tao Qin and Jianfeng Lu and Tie-Yan Liu},
      year={2020},
      eprint={2004.09297},
      archivePrefix={arXiv},
      primaryClass={cs.CL},
      url={https://arxiv.org/abs/2004.09297}, 
}

@misc{tunstall2022efficientfewshotlearningprompts,
      title={Efficient Few-Shot Learning Without Prompts}, 
      author={Lewis Tunstall and Nils Reimers and Unso Eun Seo Jo and Luke Bates and Daniel Korat and Moshe Wasserblat and Oren Pereg},
      year={2022},
      eprint={2209.11055},
      archivePrefix={arXiv},
      primaryClass={cs.CL},
      url={https://arxiv.org/abs/2209.11055}, 
}

@misc{li2020sentenceembeddingspretrainedlanguage,
      title={On the Sentence Embeddings from Pre-trained Language Models}, 
      author={Bohan Li and Hao Zhou and Junxian He and Mingxuan Wang and Yiming Yang and Lei Li},
      year={2020},
      eprint={2011.05864},
      archivePrefix={arXiv},
      primaryClass={cs.CL},
      url={https://arxiv.org/abs/2011.05864}, 
}

@misc{galinkin2024improvedlargelanguagemodel,
      title={Improved Large Language Model Jailbreak Detection via Pretrained Embeddings}, 
      author={Erick Galinkin and Martin Sablotny},
      year={2024},
      eprint={2412.01547},
      archivePrefix={arXiv},
      primaryClass={cs.CR},
      url={https://arxiv.org/abs/2412.01547}, 
}

@article{10.1109/TIT.1982.1056489,
author = {Lloyd, S.},
title = {Least squares quantization in PCM},
year = {2006},
issue_date = {March 1982},
publisher = {IEEE Press},
volume = {28},
number = {2},
issn = {0018-9448},
url = {https://doi.org/10.1109/TIT.1982.1056489},
doi = {10.1109/TIT.1982.1056489},
journal = {IEEE Trans. Inf. Theor.},
month = sep,
pages = {129–137},
numpages = {9}
}

@misc{ganguly2026trusttypical,
      title={Trust The Typical}, 
      author={Debargha Ganguly and Sreehari Sankar and Biyao Zhang and Vikash Singh and Kanan Gupta and Harshini Kavuru and Alan Luo and Weicong Chen and Warren Morningstar and Raghu Machiraju and Vipin Chaudhary},
      year={2026},
      eprint={2602.04581},
      archivePrefix={arXiv},
      primaryClass={cs.CL},
      url={https://arxiv.org/abs/2602.04581}, 
}

@misc{alanova2026crosslingualjailbreakdetectionsemantic,
      title={Cross-Lingual Jailbreak Detection via Semantic Codebooks}, 
      author={Shirin Alanova and Bogdan Minko and Sabrina Sadiekh and Evgeniy Kokuykin},
      year={2026},
      eprint={2604.25716},
      archivePrefix={arXiv},
      primaryClass={cs.CL},
      url={https://arxiv.org/abs/2604.25716}, 
}

@misc{markov2023holisticapproachundesiredcontent,
      title={A Holistic Approach to Undesired Content Detection in the Real World}, 
      author={Todor Markov and Chong Zhang and Sandhini Agarwal and Tyna Eloundou and Teddy Lee and Steven Adler and Angela Jiang and Lilian Weng},
      year={2023},
      eprint={2208.03274},
      archivePrefix={arXiv},
      primaryClass={cs.CL},
      url={https://arxiv.org/abs/2208.03274}, 
}

@inproceedings{zhan-etal-2024-injecagent,
    title = "{I}njec{A}gent: Benchmarking Indirect Prompt Injections in Tool-Integrated Large Language Model Agents",
    author = "Zhan, Qiusi  and
      Liang, Zhixiang  and
      Ying, Zifan  and
      Kang, Daniel",
    editor = "Ku, Lun-Wei  and
      Martins, Andre  and
      Srikumar, Vivek",
    booktitle = "Findings of the Association for Computational Linguistics: ACL 2024",
    month = aug,
    year = "2024",
    address = "Bangkok, Thailand",
    publisher = "Association for Computational Linguistics",
    url = "https://aclanthology.org/2024.findings-acl.624/",
    doi = "10.18653/v1/2024.findings-acl.624",
    pages = "10471--10506"
}

@inproceedings{NEURIPS2024_97091a51,
 author = {Debenedetti, Edoardo and Zhang, Jie and Balunovic, Mislav and Beurer-Kellner, Luca and Fischer, Marc and Tram\`{e}r, Florian},
 booktitle = {Advances in Neural Information Processing Systems},
 doi = {10.52202/079017-2636},
 editor = {A. Globerson and L. Mackey and D. Belgrave and A. Fan and U. Paquet and J. Tomczak and C. Zhang},
 pages = {82895--82920},
 publisher = {Curran Associates, Inc.},
 title = {AgentDojo: A Dynamic Environment to Evaluate Prompt Injection Attacks and Defenses for LLM Agents},
 volume = {37},
 year = {2024}
}

@misc{zhang2025agentsecuritybenchasb,
      title={Agent Security Bench (ASB): Formalizing and Benchmarking Attacks and Defenses in LLM-based Agents}, 
      author={Hanrong Zhang and Jingyuan Huang and Kai Mei and Yifei Yao and Zhenting Wang and Chenlu Zhan and Hongwei Wang and Yongfeng Zhang},
      year={2025},
      eprint={2410.02644},
      archivePrefix={arXiv},
      primaryClass={cs.CR},
      url={https://arxiv.org/abs/2410.02644}, 
}

@misc{gibbs2024emergingvulnerabilitiesfrontiermodels,
      title={Emerging Vulnerabilities in Frontier Models: Multi-Turn Jailbreak Attacks}, 
      author={Tom Gibbs and Ethan Kosak-Hine and George Ingebretsen and Jason Zhang and Julius Broomfield and Sara Pieri and Reihaneh Iranmanesh and Reihaneh Rabbany and Kellin Pelrine},
      year={2024},
      eprint={2409.00137},
      archivePrefix={arXiv},
      primaryClass={cs.CR},
      url={https://arxiv.org/abs/2409.00137}, 
}

\appendix

\section{Additional In-Domain Metrics}
\label{app:in_domain}

Table~\ref{tab:appendix_in_domain} reports the harmful-class F1,
false-negative rate (FNR), and false-positive rate (FPR)
for the in-domain experiments.

\begin{table}[!htbp]
\centering
\small
\setlength{\tabcolsep}{3pt}
\begin{tabular}{@{}lrrrrr@{}}
\toprule
Metric & A & B & C & D & JDPI \\
\midrule
\multicolumn{6}{@{}l}{\textbf{WildGuardMix}} \\
F1  & 94.71 & 94.85 & 94.75 & 95.01 & 94.41 \\
FNR &  6.07 &  5.57 &  6.44 &  4.78 &  6.63 \\
FPR &  4.99 &  5.31 &  4.45 &  5.91 &  5.02 \\
\midrule
\multicolumn{6}{@{}l}{\textbf{WildJailbreak}} \\
F1  & 95.68 & 95.67 & 95.69 & 95.56 & 95.39 \\
FNR &  4.51 &  3.75 &  3.88 &  3.65 &  4.68 \\
FPR &  4.42 &  5.34 &  5.13 &  5.70 &  4.85 \\
\midrule
\multicolumn{6}{@{}l}{\textbf{BeaverTailsQA}} \\
F1  & 83.56 & 83.67 & 83.71 & 83.81 & 81.37 \\
FNR & 18.57 & 16.88 & 16.80 & 16.99 & 23.08 \\
FPR & 16.75 & 19.34 & 19.37 & 18.73 & 15.10 \\
\midrule
\multicolumn{6}{@{}l}{\textbf{TomGibbs}} \\
F1  & 96.39 & 96.56 & 96.38 & 96.49 & 91.76 \\
FNR &  1.23 &  1.23 &  1.58 &  0.70 &  9.86 \\
FPR & 10.67 & 10.06 & 10.06 & 11.28 & 10.98 \\
\bottomrule
\end{tabular}
\caption{Additional in-domain metrics for Methods A--E.
All values are percentages.}
\label{tab:appendix_in_domain}
\end{table}

\begin{figure}[t]
\centering

\includegraphics[width=\columnwidth]
{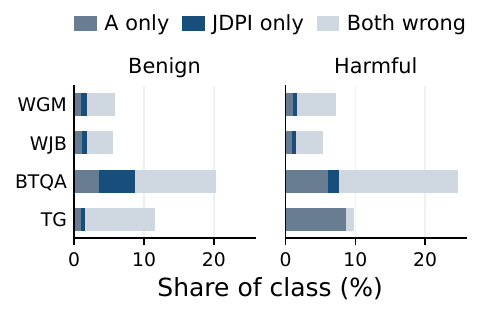}

\caption{
Error composition of Methods A and E across four benchmarks,
separated by the true benign and harmful classes.
Bars show the percentages of examples for which only A is correct,
only E is correct, or both methods are wrong.
Percentages are computed within each true class;
examples correctly classified by both methods are omitted.
}

\label{fig:ae_disagreement}
\end{figure}

Figure~\ref{fig:ae_disagreement} further shows that the
two methods make different errors.
Examples correctly classified only by JDPI demonstrate
that prototype inference does not simply reproduce
the classifier's predictions.
Conversely, examples correctly classified only by A
show where the learned decision function remains useful.

\subsection{Effect of Contrastive Fine-Tuning}
\label{app:finetuning_effect}

We compare the original encoder with the fine-tuned
encoder used for each benchmark's main results to assess the effect of contrastive adaptation. Appendix~\ref{app:in_domain} reports additional in-domain metrics. The subsequent analyses focus on Method~A and JDPI to assess how much detection performance direct prototype inference retains without a trained downstream classifier.

\begin{figure}[t]
\centering
\includegraphics[width=\columnwidth]
{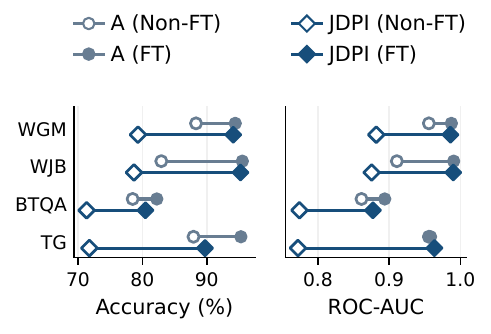}
\caption{Effect of contrastive fine-tuning on Methods A and JDPI. Open and filled markers denote the original and fine-tuned encoders, respectively.}
\label{fig:finetuning_effect}
\end{figure}

The results are summarized in
Figure~\ref{fig:finetuning_effect}.
For TomGibbs, the comparison uses the final Phase~2 checkpoint and therefore reflects the combined effect of both training phases.

As shown in Figure \ref{fig:finetuning_effect}, the fine-tuned embedding model benefits not only prototype-based detection but also produces a clear improvement for the baseline classifier. More importantly, after contrastive learning, Methods A and E achieve comparable accuracy, particularly on WildJailbreak and WildGuardMix.

Contrastive fine-tuning improves JDPI across all four benchmarks. Its accuracy increases by 14.81 percentage points on WildGuardMix, 16.54 on WildJailbreak, 9.14 on BeaverTailsQA, and 17.97 on TomGibbs. Classifier~A also improves, with accuracy gains ranging from 3.75 to 12.59 percentage points.

The larger gains for JDPI suggest that contrastive adaptation makes the representation geometry more useful for direct prototype-based decisions.
In particular, E approaches A on WildGuardMix and WildJailbreak after fine-tuning. The resulting performance therefore depends on the combination of representation learning and prototype inference, rather than on prototype construction alone.

\section{Complete Cross-Dataset Results}
\label{app:cross_dataset}

Tables~\ref{tab:appendix_transfer_performance}
and~\ref{tab:appendix_transfer_errors} report results for
classifier~A, variants~B--D, and JDPI across all 12 directed
transfers. WGM, WJB, BTQA, and TG denote WildGuardMix,
WildJailbreak, BeaverTailsQA, and TomGibbs, respectively.
Each mean gives equal weight to each transfer direction.

\begin{table*}[t]
\centering
\small
\setlength{\tabcolsep}{3pt}
\begin{tabular}{@{}l*{10}{r}@{}}
\toprule
& \multicolumn{5}{c}{Accuracy (\%)}
& \multicolumn{5}{c}{ROC-AUC} \\
\cmidrule(lr){2-6}\cmidrule(lr){7-11}
Source $\to$ Target & A & B & C & D & JDPI & A & B & C & D & JDPI \\
\midrule
WGM $\to$ WJB  & 92.48 & 92.69 & 92.07 & 93.04 & 91.97 & 0.9812 & 0.9813 & 0.9816 & 0.9820 & 0.9797 \\
WGM $\to$ BTQA & 68.26 & 67.88 & 67.98 & 67.82 & 68.54 & 0.7406 & 0.7407 & 0.7393 & 0.7394 & 0.7439 \\
WGM $\to$ TG   & 77.57 & 78.79 & 78.35 & 78.68 & 77.46 & 0.8409 & 0.8461 & 0.8422 & 0.8422 & 0.8307 \\
\addlinespace
WJB $\to$ WGM  & 90.58 & 90.43 & 90.79 & 89.88 & 90.91 & 0.9699 & 0.9708 & 0.9705 & 0.9690 & 0.9685 \\
WJB $\to$ BTQA & 58.30 & 58.57 & 58.61 & 57.89 & 61.09 & 0.5952 & 0.5982 & 0.6063 & 0.5950 & 0.6345 \\
WJB $\to$ TG   & 72.88 & 72.10 & 72.43 & 70.76 & 74.00 & 0.7689 & 0.7669 & 0.7660 & 0.7605 & 0.7784 \\
\addlinespace
BTQA $\to$ WGM & 74.27 & 73.68 & 74.86 & 71.31 & 73.34 & 0.8030 & 0.8180 & 0.8185 & 0.8030 & 0.8138 \\
BTQA $\to$ WJB & 68.38 & 69.80 & 69.81 & 69.69 & 63.65 & 0.7620 & 0.7642 & 0.7647 & 0.7657 & 0.7276 \\
BTQA $\to$ TG  & 66.63 & 66.07 & 66.74 & 65.62 & 70.42 & 0.6888 & 0.6910 & 0.6937 & 0.6919 & 0.7227 \\
\addlinespace
TG $\to$ WGM   & 57.26 & 57.65 & 58.83 & 57.59 & 66.85 & 0.6896 & 0.6911 & 0.7011 & 0.6787 & 0.7135 \\
TG $\to$ WJB   & 59.65 & 59.67 & 60.21 & 59.31 & 61.33 & 0.6660 & 0.6618 & 0.6631 & 0.6612 & 0.6541 \\
TG $\to$ BTQA  & 57.75 & 57.89 & 58.67 & 57.86 & 63.49 & 0.6604 & 0.6606 & 0.6628 & 0.6544 & 0.6655 \\
\midrule
Mean & 70.33 & 70.44 & 70.78 & 69.96 & 71.92
     & 0.7639 & 0.7659 & 0.7675 & 0.7619 & 0.7694 \\
\bottomrule
\end{tabular}
\caption{Cross-dataset accuracy and ROC-AUC for all five methods.}
\label{tab:appendix_transfer_performance}
\end{table*}

\begin{table*}[t]
\centering
\small
\setlength{\tabcolsep}{3pt}
\begin{tabular}{@{}l*{10}{r}@{}}
\toprule
& \multicolumn{5}{c}{FNR (\%)}
& \multicolumn{5}{c}{FPR (\%)} \\
\cmidrule(lr){2-6}\cmidrule(lr){7-11}
Source $\to$ Target & A & B & C & D & JDPI & A & B & C & D & JDPI \\
\midrule
WGM $\to$ WJB  & 11.07 & 10.27 & 12.05 &  9.03 & 11.73 &  3.71 &  4.14 &  3.51 &  4.75 &  4.06 \\
WGM $\to$ BTQA & 20.40 & 19.30 & 21.46 & 18.57 & 22.33 & 45.83 & 48.06 & 45.14 & 49.11 & 42.81 \\
WGM $\to$ TG   & 11.44 &  8.45 & 11.27 &  7.04 & 11.27 & 41.46 & 43.29 & 39.63 & 46.04 & 42.07 \\
\addlinespace
WJB $\to$ WGM  &  6.49 &  5.80 &  6.19 &  5.17 &  5.77 & 12.74 & 13.83 & 12.63 & 15.72 & 12.84 \\
WJB $\to$ BTQA & 37.20 & 33.97 & 37.09 & 31.87 & 34.94 & 47.29 & 50.70 & 46.74 & 54.86 & 43.85 \\
WJB $\to$ TG   & 15.32 & 13.38 & 12.85 & 13.56 & 17.78 & 47.56 & 53.05 & 53.05 & 56.40 & 40.24 \\
\addlinespace
BTQA $\to$ WGM & 28.22 & 18.04 & 22.24 & 16.51 & 38.96 & 22.91 & 35.70 & 28.43 & 42.47 & 12.74 \\
BTQA $\to$ WJB & 44.04 & 33.90 & 32.61 & 32.68 & 58.34 & 18.29 & 26.23 & 27.59 & 27.77 & 12.74 \\
BTQA $\to$ TG  &  4.75 &  3.52 &  4.05 &  4.58 &  9.86 & 82.93 & 86.59 & 83.84 & 85.98 & 63.72 \\
\addlinespace
TG $\to$ WGM   &  8.15 &  8.66 & 10.46 &  8.27 & 31.85 & 81.90 & 80.49 & 75.94 & 81.07 & 34.62 \\
TG $\to$ WJB   & 10.84 & 11.62 & 13.29 & 10.25 & 37.67 & 72.03 & 71.14 & 68.23 & 73.36 & 39.75 \\
TG $\to$ BTQA  &  2.94 &  3.41 &  4.49 &  3.43 & 32.04 & 91.11 & 90.23 & 87.13 & 90.26 & 42.06 \\
\midrule
Mean & 16.74 & 14.19 & 15.67 & 13.41 & 26.05
     & 47.31 & 50.29 & 47.65 & 52.31 & 32.63 \\
\bottomrule
\end{tabular}
\caption{Cross-dataset false-negative and false-positive rates
for all five methods.}
\label{tab:appendix_transfer_errors}
\end{table*}

The two transfer directions between WildGuardMix and WildJailbreak
contain shared prompt texts between source development data and
target test data. Excluding both directions, JDPI achieves higher
accuracy than classifier~A in 7 of the remaining 10 directions,
with a mean accuracy advantage of 1.92 percentage points across
all 10 directions.

\section{Implementation Details}
\label{app:implementation}

\begin{table}[t]
\centering
\small
\setlength{\tabcolsep}{3pt}
\begin{tabular}{@{}lrrrr@{}}
\toprule
Dataset & A & JDPI & Saved & Reduction \\
\midrule
WildGuardMix  & 51,846 & 15,846 & 36,000 & 69.4\% \\
WildJailbreak & 47,994 & 11,994 & 36,000 & 75.0\% \\
BeaverTailsQA & 51,846 & 15,846 & 36,000 & 69.4\% \\
TomGibbs      &  4,368 &  1,488 &  2,880 & 65.9\% \\
\bottomrule
\end{tabular}
\caption{Labeled examples used for model development.
Both methods use the structure and validation splits.
Classifier~A also uses a separate classifier-training split.
Counts are based on dataset rows and exclude the test split.
Reduction is relative to classifier~A.}
\label{tab:label_budget}
\end{table}

\paragraph{Encoder adaptation.}
Table~\ref{tab:encoder_training} reports the settings of the
saved encoders. The training batch size was 8 in all runs
(per device for WildJailbreak). WGM, WJB, BTQA, and TG denote
WildGuardMix, WildJailbreak, BeaverTailsQA, and TomGibbs,
respectively.

\begin{table}[t]
\centering
\small
\setlength{\tabcolsep}{3pt}
\begin{tabular}{@{}llrrrr@{}}
\toprule
Data & Run & Pairs & Epochs & Warm-up & LR \\
\midrule
WGM  & Phase 1 & $\sim$16,000 & 5 & 200 & $2{\times}10^{-5}$ \\
BTQA & Phase 1 & $\sim$16,000 & 5 & 200 & $2{\times}10^{-5}$ \\
TG   & Phase 1 & 16,000       & 5 & 200 & $2{\times}10^{-5}$ \\
WJB & Phase 1 & 16,000 & 5 & 200 & $2{\times}10^{-5}$ \\
\midrule
WGM  & Phase 2 & $\sim$6,000  & 2 & 50  & $2{\times}10^{-5}$ \\
BTQA & Phase 2 & $\sim$6,000  & 2 & 50  & $2{\times}10^{-5}$ \\
TG   & Phase 2 & 5,000        & 2 & 50  & $2{\times}10^{-5}$ \\
\bottomrule
\end{tabular}
\caption{Encoder training settings. Phase~2 values are per
refinement iteration. Warm-up is measured in optimizer steps.}
\label{tab:encoder_training}
\end{table}

\paragraph{Method D configuration.}
We fixed the benign class weight at 1.0 and tested harmful class weights of 1.0, 1.2, and 1.5. For each weight, we searched thresholds in $\{0.10,0.15,\ldots,0.90\}$ on the
validation split. Among thresholds with precision of at least 0.70 and an FPR no more than one percentage point above Method B's reference FPR, we selected the threshold with
the highest recall. If no threshold met both constraints, we selected by recall over the full grid. Recall ties were broken by F1, balanced accuracy, and accuracy, in that order.

We computed Method~B's reference FPR and ranked the candidate class weights on the validation split. The ranking prioritized fewer false negatives, followed by lower FPR.

\section{Ablation Study: Harmful-Subcluster Refinement}
\label{app:refinement}

In the second phase, we further refine the internal structure of the harmful representations through an iterative clustering procedure. At each iteration, harmful samples are re-encoded using the current encoder and partitioned using $K$-means clustering. The number of clusters is selected from $K=4,\ldots,12$ by maximizing the silhouette score computed using cosine distance.
The resulting cluster assignments are used as pseudo-labels to construct new contrastive pairs. Harmful samples belonging to the same cluster form positive pairs, while harmful samples from different clusters form negative pairs. Harmful-benign pairs are additionally retained as negative pairs to preserve the global separation between the two classes.

Using these newly constructed pairs, the encoder is further optimized with the same CoSENTLoss. The updated encoder is then used to re-encode harmful samples, and the clustering process is repeated for up to a predefined maximum number of iterations.
By progressively refining substructures within the harmful class, this stage allows us to examine whether finer cluster separation improves binary harmful--benign detection.

\begin{figure}[t]
\centering
\includegraphics[width=\columnwidth]
{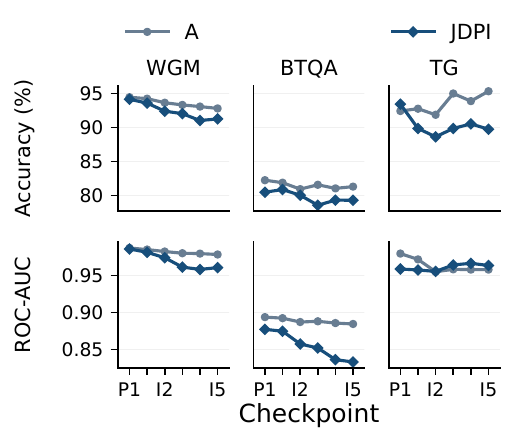}
\caption{Accuracy and ROC-AUC across refinement
checkpoints for Methods A and JDPI.}
\label{fig:checkpoint_performance}
\end{figure}

\paragraph{WildGuardMix and BeaverTailsQA.}
Table~\ref{tab:checkpoint_geometry} reports the silhouette and detection results across refinement checkpoints for WildGuardMix and BeaverTailsQA. This analysis focuses on the performance gap between JDPI and the classifier-based baselines during iterative representation refinement. The refinement analysis is conducted on WildGuardMix and BeaverTails QA. The checkpoint results are shown in Figure~\ref{fig:checkpoint_performance}.

\begin{table}[!htbp]
\centering
\small
\setlength{\tabcolsep}{2.5pt}
\begin{tabular}{@{}lrrrrrr@{}}
\toprule
& \multicolumn{2}{c}{Silhouette}
& \multicolumn{2}{c}{Acc. (\%)}
& \multicolumn{2}{c}{ROC-AUC} \\
\cmidrule(lr){2-3}\cmidrule(lr){4-5}\cmidrule(lr){6-7}
Stage & Global & Harm. & A & E & A & JDPI \\
\midrule
\multicolumn{7}{@{}l}{\textbf{WildGuardMix}} \\
P1 & 0.9065 & 0.2209 & 94.43 & 94.13 & 0.9874 & 0.9860 \\
I1 & 0.6173 & 0.7029 & 94.22 & 93.57 & 0.9852 & 0.9815 \\
I2 & 0.5389 & 0.8176 & 93.62 & 92.38 & 0.9826 & 0.9742 \\
I3 & 0.5217 & 0.8627 & 93.30 & 92.02 & 0.9803 & 0.9615 \\
I4 & 0.5147 & 0.8762 & 93.06 & 91.01 & 0.9798 & 0.9584 \\
I5 & 0.4691 & 0.8875 & 92.79 & 91.24 & 0.9786 & 0.9608 \\
\midrule
\multicolumn{7}{@{}l}{\textbf{BeaverTailsQA}} \\
P1 & 0.8030 & 0.1221 & 82.24 & 80.48 & 0.8938 & 0.8772 \\
I1 & 0.5993 & 0.3132 & 81.87 & 80.87 & 0.8923 & 0.8747 \\
I2 & 0.5654 & 0.5454 & 80.93 & 80.03 & 0.8871 & 0.8574 \\
I3 & 0.5227 & 0.6395 & 81.57 & 78.56 & 0.8881 & 0.8519 \\
I4 & 0.4420 & 0.6816 & 81.04 & 79.31 & 0.8857 & 0.8361 \\
I5 & 0.4507 & 0.7247 & 81.30 & 79.30 & 0.8845 & 0.8330 \\
\bottomrule
\end{tabular}
\caption{Refinement checkpoint results. Global denotes
the harmful--benign silhouette and Harm.\ denotes
the silhouette among harmful subclusters. P1 is the
Phase~1 checkpoint and I1--I5 are the Phase~2 iterations.}
\label{tab:checkpoint_geometry}
\end{table}

Refinement increases harmful-cluster separation while
reducing global harmful--benign separation on both
benchmarks. On WildGuardMix, the harmful-cluster silhouette rises
from 0.2209 to 0.8875, whereas the global silhouette
falls from 0.9065 to 0.4691. On BeaverTailsQA, the corresponding changes are
0.1221 to 0.7247 and 0.8030 to 0.4507. 

Over the same checkpoints, JDPI's ROC-AUC decreases from
0.9860 to 0.9608 on WildGuardMix and from 0.8772 to
0.8330 on BeaverTailsQA.
A is less affected, with ROC-AUC decreasing from
0.9874 to 0.9786 and from 0.8938 to 0.8845,
respectively.
Thus, improved separation among harmful clusters
does not translate into improved binary detection
in these experiments.

These trends are consistent with a mismatch between
the refinement and detection objectives.
Phase~1 explicitly separates harmful and benign
examples, whereas Phase~2 additionally separates
subgroups within the harmful class.
This can alter the geometry used by nearest-prototype
inference without improving the harmful--benign
decision boundary.
A learned classifier may accommodate these changes
more effectively by fitting a decision direction
across embedding dimensions. The results therefore favor evaluating refinement
through downstream detection performance, rather than harmful-cluster silhouette alone.

Future work could jointly optimize harmful--benign separation and harmful-subcluster separation, with their relative weights selected on validation data for the intended detection protocol.

\paragraph{TomGibbs.}
We evaluate Phase~1 and Iterations~1--5 on the same 896 held-out conversations. For multi-turn replay, we use each checkpoint's thresholds from complete-prompt validation.

\begin{table}[t]
\centering
\small
\setlength{\tabcolsep}{3pt}
\begin{tabular}{@{}lrrrrr@{}}
\toprule
Checkpoint & A & B & C & D & JDPI \\
\midrule
\multicolumn{6}{@{}l}{\textit{Accuracy (\%)}} \\
Phase 1  & 92.41 & 92.08 & 92.52 & 91.63 & 93.42 \\
Iter.\ 1 & 92.75 & 93.75 & 93.86 & 94.20 & 89.84 \\
Iter.\ 2 & 91.85 & 92.19 & 93.53 & 93.97 & 88.62 \\
Iter.\ 3 & 94.98 & 95.42 & 95.31 & 95.42 & 89.84 \\
Iter.\ 4 & 93.86 & 93.08 & 94.08 & 93.08 & 90.51 \\
Iter.\ 5 & 95.31 & 95.54 & 95.31 & 95.42 & 89.73 \\
\midrule
\multicolumn{6}{@{}l}{\textit{ROC-AUC}} \\
Phase 1  & .9798 & .9760 & .9661 & .9671 & .9589 \\
Iter.\ 1 & .9719 & .9702 & .9706 & .9744 & .9576 \\
Iter.\ 2 & .9552 & .9557 & .9592 & .9674 & .9557 \\
Iter.\ 3 & .9585 & .9575 & .9559 & .9564 & .9644 \\
Iter.\ 4 & .9581 & .9586 & .9565 & .9584 & .9666 \\
Iter.\ 5 & .9581 & .9586 & .9566 & .9587 & .9637 \\
\bottomrule
\end{tabular}
\caption{Complete-prompt detection across TomGibbs checkpoints.}
\label{tab:tomgibbs_static_checkpoints}
\end{table}

\begin{table}[t]
\centering
\small
\setlength{\tabcolsep}{2.5pt}
\begin{tabular}{@{}lrrrrrr@{}}
\toprule
& \multicolumn{3}{c}{Method A} & \multicolumn{3}{c}{JDPI} \\
\cmidrule(lr){2-4}\cmidrule(lr){5-7}
Checkpoint & Acc. & FNR & FPR & Acc. & FNR & FPR \\
\midrule
\multicolumn{7}{@{}l}{\textit{Current turn}} \\
Phase 1  & 63.39 & 0.00 & 100.00 & 63.39 & 0.00  & 100.00 \\
Iter.\ 1 & 63.39 & 0.00 & 100.00 & 76.67 & 2.82  & 58.84 \\
Iter.\ 2 & 63.39 & 0.00 & 100.00 & 78.79 & 17.25 & 28.05 \\
Iter.\ 3 & 63.39 & 0.00 & 100.00 & 75.00 & 6.34  & 57.32 \\
Iter.\ 4 & 70.54 & 0.00 & 80.49  & 74.55 & 1.41  & 67.07 \\
Iter.\ 5 & 63.39 & 0.00 & 100.00 & 77.23 & 6.34  & 51.22 \\
\midrule
\multicolumn{7}{@{}l}{\textit{Accumulated user history}} \\
Phase 1  & 63.39 & 0.00 & 100.00 & 63.39 & 0.00 & 100.00 \\
Iter.\ 1 & 63.39 & 0.00 & 100.00 & 86.94 & 4.93 & 27.13 \\
Iter.\ 2 & 63.39 & 0.00 & 100.00 & 87.72 & 8.80 & 18.29 \\
Iter.\ 3 & 63.39 & 0.00 & 100.00 & 87.61 & 5.46 & 24.39 \\
Iter.\ 4 & 87.61 & 1.94 & 30.49  & 83.48 & 4.23 & 37.80 \\
Iter.\ 5 & 63.39 & 0.00 & 100.00 & 89.29 & 4.23 & 21.95 \\
\bottomrule
\end{tabular}
\caption{Conversation-level any-turn results across TomGibbs
checkpoints. All values are percentages.}
\label{tab:tomgibbs_replay_checkpoints}
\end{table}

JDPI's complete-prompt accuracy is highest at Phase~1, and its
ROC-AUC peaks at Iteration~4
(Table~\ref{tab:tomgibbs_static_checkpoints}).
With accumulated user history, its any-turn accuracy improves
after the first refinement iteration and varies across subsequent
checkpoints, reaching 89.29\% at Iteration~5
(Table~\ref{tab:tomgibbs_replay_checkpoints}).
Method~A leads JDPI at Iteration~4 under this replay setting.
From Phase~1 to Iteration~5, JDPI's validation-selected threshold
changes from 0.25 to 0.60. With accumulated user history, Its replay false-positive rate falls from 100.00\% to 21.95\%, and its false-negative rate increases
from 0.00\% to 4.23\%.

\section{Additional Multi-Turn Analysis}
\label{app:multiturn_subgroups}

We report conversation-level false-positive rates separately
for completely benign and semi-benign conversations on
TomGibbs under the any-turn rule.

\begin{table}[t]
\centering
\small
\setlength{\tabcolsep}{4pt}
\begin{tabular}{@{}llrr@{}}
\toprule
Benign subgroup & Input & A & JDPI \\
\midrule
Completely benign ($n=168$) & Current & 100.00 & 28.57 \\
                            & History & 100.00 &  0.00 \\
Semi-benign ($n=160$)       & Current & 100.00 & 75.00 \\
                            & History & 100.00 & 45.00 \\
\bottomrule
\end{tabular}
\caption{Any-turn false-positive rates on TomGibbs (\%).}
\label{tab:tomgibbs_benign_subgroups}
\end{table}

\end{document}